\documentclass{
article}

\usepackage[main, final]{neurips_2026}

\usepackage[utf8]{inputenc} 
\usepackage[T1]{fontenc}    
\usepackage{hyperref}       
\usepackage{url}            
\usepackage{booktabs}       
\usepackage{amsfonts}       
\usepackage{nicefrac}       
\usepackage{microtype}      
\usepackage{xcolor}         

\usepackage{comment}
\usepackage{amsmath}
\usepackage{lineno}
\usepackage{amssymb}
\usepackage{mathtools}
\usepackage[inline]{enumitem}
\usepackage{amsthm}
\usepackage{subcaption}
\usepackage{wrapfig}
\usepackage[capitalize,noabbrev]{cleveref}
\usepackage{float}
\usepackage{pgfplots}
\usepackage{wrapfig}

\usepgfplotslibrary{groupplots}
\pgfplotsset{compat=1.18}
\PassOptionsToPackage{numbers}{natbib}
\usetikzlibrary{external}
\usepackage{sansmath}
\usepackage{etoolbox} 
\usepackage{makecell}
\usepackage{multirow}
\usepackage{subcaption}
\usepackage{placeins}
\usepackage{acro}
\usepackage{cleveref}
\usepackage{etoc}

\makeatletter
\newcommand{\AppendixTOCStyle}{%
  \etocsetstyle{section}{}{}
    {\l@section
       {\large\mdseries\numberline{\etocnumber}\scshape\etocname}
       {\mdseries\etocpage}}
    {}%
  \etocsetstyle{subsection}{}{}
    {\l@subsection
       {\numberline{\etocnumber}\scshape\etocname}
       {\etocpage}}
    {}%
}
\makeatother

\definecolor{s13}{cmyk}{.75,.50,.0,.0} 
\hypersetup{
    pdfdisplaydoctitle,
    bookmarksnumbered=true,
    bookmarksopen,
    breaklinks,
    linktoc=page,                     
    plainpages=false,
    unicode=true,
    colorlinks=true,
    allcolors=blue!50!black,             
    urlcolor=blue!50!black,                     
}
\setcitestyle{round}

\DeclareAcronym{all}{
  short="All",
  sort=all,
  long=variance head on all latent representations,
  first-style=short,
  tag=nolist
}
\DeclareAcronym{ensemble}{
  short=Ens.,
  sort=ens,
  long=ensemble of variance heads,
  first-style=short,
  tag=nolist
}
\DeclareAcronym{scp}{
  short=SCP,
  sort=scp,
  long=split conformal prediction
}
\DeclareAcronym{lacp}{
  short=LACP,
  sort=lacp,
  long=locally adaptive conformal prediction
}
\DeclareAcronym{mlp}{
  short=MLP,
  sort=mlp,
  long=multilayer perceptron
}
\DeclareAcronym{nll}{
  short=NLL,
  sort=nll,
  long=negative log likelihood
}

\title{Deep Heteroskedastic Regression: Post-Hoc Variance Estimation from Latent Representations}

\author{%
  Mikkel Jordahn\thanks{Equal contribution}\\
  Technical University of Denmark\\
  \texttt{mikkjo@dtu.dk}
  \And
  Jonas Vestergaard Jensen\footnotemark[1] \\
  Technical University of Denmark\\
  \texttt{jovje@dtu.dk}
  \And
  James Harrison\\
  Google DeepMind\\
  \texttt{jamesharrison@google.com}
  \And
  Michael Riis Andersen\\
  Technical University of Denmark\\
  \texttt{miri@dtu.dk}
  \And
  Mikkel N.~Schmidt\\
  Technical University of Denmark\\
  \texttt{mnsc@dtu.dk}
}

\begin{document}

\maketitle
\etocdepthtag.toc{mtmain}

\begin{abstract}
    Uncertainty quantification (UQ) in deep learning regression is of wide interest, as it supports critical applications including sequential decision making and risk-sensitive tasks. In heteroskedastic regression, where the uncertainty of the target depends on the input, a common approach is to train a neural network that parametrises the mean and the variance of the predictive distribution. Yet to this day, training deep heteroskedastic regression models poses severe practical challenges in the trade-off between uncertainty quantification and mean prediction, such as variance overfitting, optimization difficulties and representation collapse. In this work, we identify and discuss the core issues and propose a simple and efficient procedure that addresses these jointly by post-hoc fitting a variance model across the intermediate layers of a pretrained network on a hold-out dataset. We demonstrate that this method is competitive with end-to-end trained mean-variance networks in heteroskedastic UQ on several data modalities. The method retains mean prediction accuracy, is cheap at both train and prediction time, and requires no additional data compared to existing methods. Finally, we show this method works on large scale foundation models for chemistry, paving the way for cheap heteroskedastic UQ in large scale neural networks without having to retrain such models end-to-end.
\end{abstract}

\section{Introduction}

Uncertainty quantification (UQ) is of wide interest in the field of deep learning due to its utility in several applications, ranging from sequential decision making to safety-critical use-cases \citep{springenberg2016bayesian,chitta2019largescalevisualactivelearning,li2024study,brunzema2024bayesian, cinquin2025what,duran-martin2025martingale}. One approach is deep heteroskedastic regression~\citep{nix1994estimating}, commonly referred to as mean-variance networks, where a neural network predicts an input-dependent variance along with the mean. This variance absorbs the total predictive residual error, treating it  as aleatoric uncertainty~\citep{Gal2016UncertaintyID}. In this work, we concern ourselves with learning input-dependent uncertainty directly as a neural network output while maintaining scalability, low prediction cost, and avoiding significant increases in training time or degradation in mean prediction accuracy. In particular, we consider the overparameterized neural network regime, and focus on settings relevant to foundation models.

In recent years, several works have focused on improving deep heteroskedastic regression \citep{skafte2019reliable,stirn2020variational,seitzer2022pitfalls,stirn2023faithful,wong-toi2024understanding, immer2024effective, harrison2025heteroscedastic}, and generally the methods follow the recipe originally proposed by~\citet{nix1994estimating}, with modifications during training. One or more neural networks are trained to output the parameters of some predictive distribution (most commonly, a Gaussian distribution), using negative log-likelihood as the training criterion. Common for many of these works is that they claim that existing training methods either result in loss of mean precision or poor variance estimates. Some argue that the precision loss is due to optimisation issues of the Gaussian negative-log likelihood \citep{skafte2019reliable,seitzer2022pitfalls,stirn2023faithful,immer2024effective} or that an accurate variance estimate requires additional mean regularisation \citep{wong-toi2024understanding}. Interestingly, throughout our experiments, we find that many of the existing methods perform strongly across a number of data modalities and neural network architectures if allowed to train until mean convergence, provided that a hold-out dataset is used to recalibrate their variance estimates after training. Without such hold-out datasets, the scale of the variance estimates is shrunk to near-zero, leading to extremely poor test-time uncertainty quantification. This leads us to ask: 
\textit{Can a post-hoc variance estimation procedure applied to a conventionally trained modern and deep mean-only network achieve uncertainty estimation performance comparable to that of end-to-end optimized mean–variance networks?}

To design such a method, we first discuss three core problems with training heteroskedastic regression neural networks, some of which have been discussed in previous works but have not, to the best of our knowledge, been addressed jointly \citep{nix1994learning,skafte2019reliable,stirn2020variational,seitzer2022pitfalls,stirn2023faithful,wong-toi2024understanding, immer2024effective, harrison2025heteroscedastic}. 
Secondly, we devise a simple and effective post-hoc variance estimation method using linear heads on intermediate layers that incurs minimal additional training and inference cost, and show that it performs at least on par with end-to-end trained mean–variance networks.
Thirdly, we conduct several ablation studies illustrating the efficacy and requirements for our method, and we demonstrate its applicability through large-scale machine learning interatomic potentials as a representative use case \citep{qu2024importance,UMA}.

To summarize, the contributions of this work are the following:

\begin{enumerate}[label=(\roman*),topsep=0.05em, itemsep=0.05em]
    \item We identify and characterize three core problems in deep heteroskedastic regression.
    \item We revisit post-hoc variance fitting for neural networks and device a method for deep networks that requires no train-time tuning, incurs minimal inference cost, and requires no additional data compared to existing methods.
    \item We provide a wide range of ablation studies, showcasing that our variance estimator design is necessarily simple yet robust.
    \item We showcase that our method scales and performs well on large scale foundation models.
\end{enumerate}

\section{Deep Heteroskedastic Regression Pitfalls}
\label{sec:pitfalls}

\paragraph{Preliminaries} In the regression setting, we generally have an i.i.d. dataset $\mathcal{D}= \{(x_i, y_i)\}^N_{i=1}$ where $x \in \mathbb{R}^d$ is the input and $y \in \mathbb{R}$ is the corresponding output . From this dataset, we want to learn a function estimator $f_\theta$, which we can use to predict future $y^*$ given a new $x^*$, i.e., $y^*=f_\theta(x^*)$. In many applications, it is of interest to have an uncertainty estimate on the predicted $y^*$, and in this work, we study the \textit{heteroskedastic} setting, meaning that the noise variance is input-dependent. Most commonly, this is done using a Gaussian predictive distribution, such that $p_\theta(y^*|x^*)=\mathcal{N}(y^*|\mu_\theta(x^*), \sigma_\theta^2(x^*))$. We will refer to neural networks outputting these parameters as \textit{mean-variance networks}. Importantly, throughout this work we assume that we are in the over-parametrized neural network regime, as is generally the case in deep learning applications today.

In the following, we further elaborate on the three identified problems and the existing literature in deep heteroskedastic regression on solutions to these when applicable. The three problems we discuss are: variance overfitting, optimisation issues, and representation related issues in modern deep neural networks. Whilst we generally do not observe the optimisation issues described in \cref{sec:opt_issues} in our experiments, we include the optimization aspect here for completeness as it has been the topic of many previous works on mean-variance networks.

\subsection{Variance Overfitting}
\label{sec:underestimated_residuals}
The first issue we consider is variance overfitting. This mechanism appears in the overparametrized neural network regime where the network has sufficient capacity to interpolate all training data points perfectly, even in the presence of noise. When this happens, the model variances $\sigma^2_\theta(x)$ approach zero for all  training points. This training dynamic can be confirmed by inspecting the Gaussian negative log-likelihood (NLL) and its gradients:

\begin{equation}
    \label{eq:nll_loss}
    \mathcal{L}(\theta, \mathcal{D})=\frac{1}{2}\sum_{(x,y)\in \mathcal{D}}\left[\log\sigma_\theta^2(x)+\frac{(y-\mu_\theta(x))^2}{\sigma^2_\theta(x)}\right],
\end{equation}

\begin{align}
    \nabla_{\mu_\theta(x)}\mathcal{L}=\sum_{(x,y)\in \mathcal{D}}\frac{y-\mu_\theta(x)}{\sigma^2_\theta(x)}, &  &
    \nabla_{\sigma^2_\theta(x)}\mathcal{L}=\frac{1}{2}\sum_{(x,y)\in \mathcal{D}}\frac{\sigma_\theta^2(x)-(y-\mu_\theta(x))^2}{\sigma^4_\theta(x)}.
\end{align}

As training progresses, residuals $r(x)=(y-\mu_\theta(x))\rightarrow0$, and in turn, the only remaining term in the variance gradient is strictly positive $\frac{1}{\sigma^2_\theta(x)}>0$, which will drive the predicted variances to zero at an accelerating rate as the variance shrinks.

Other works have proposed regularization of the learned mean and variance function as a solution to this problem \citep{wong-toi2024understanding}, but this requires hyperparameter tuning that often is not feasible for large-scale models. Moreover, it is known from the double descent literature that in the highly overparameterized regime, generalization error improves even after training points are perfectly interpolated \citep{belkin2019reconciling, nakkiran2019deepdoubledescentbigger, belik2020two}, which leads us to question if regularization of the mean is the right avenue to resolve this problem. Instead, we hypothesize that using hold-out data on which the mean has not been fit, to fit the variance estimator, will provide better results by directly bounding the variance collapse, as also argued by \citet{nix1994learning} for shallow networks.

\subsection{Optimisation Issues}
\label{sec:opt_issues}
The second problem concerns optimization difficulties that others have observed for gradient-based maximum likelihood-methods. This phenomenon is generally presented by again inspecting the gradients of the Gaussian negative log-likelihood (NLL) with respect to $\mu_\theta(x)$ and $\sigma_\theta^2(x)$. Others have argued that if the predicted variance for a datapoint $x$ grows large during training, the updates to both $\mu_\theta(x)$ and $\sigma_\theta^2(x)$ slow down. Previous works argue that this often causes mean-variance networks trained this way to produce inaccurate mean predictions and inflated variances. For more details, we refer to \citet{seitzer2022pitfalls}, who present a thorough analysis of the general optimisation issues encountered when training mean-variance networks end-to-end.

Several works have discussed possible solutions to this problem. \citet{seitzer2022pitfalls} propose $\beta$-NLL which applies a variance-reweighting term $\sigma^{2\beta}_\theta(x)$ to the NLL, where $\beta$ is a hyperparameter. Similar to the phased training of \citet{nix1994learning}, \citet{skafte2019reliable} propose to first train only the mean of the network, and then move to \textit{mean-variance split training} where they alternate between optimizing $\mu_\theta(x)$ and $\sigma^2_\theta(x)$, which requires tuning the warm-up period for $\mu_\theta(x)$. \citet{takahashi2018student} and \citet{skafte2019reliable} both argue that a Student's t parametrization stabilizes these optimisation issues. Finally, a number of works argue for Bayesian approaches, in which a prior is placed on the predictive variance \citep{stirn2020variational, harrison2025heteroscedastic}. Common to most of these methods is that they introduce hyperparameters that require additional tuning, and empirically often come at the cost of degraded mean estimation.

To our knowledge, the \emph{faithful} approach of \citet{stirn2023faithful} is the only method that preserves mean estimation while resolving these optimisation issues without additional tuning cost for end-to-end trained mean-variance netowrks. This is achieved by detaching the gradients of the variance predictor from the computational graph, such that the uncertainty prediction does not affect the learning of the backbone of the network, and by rescaling the gradient of $\mu_\theta(x)$ by $\sigma^2_\theta$ such that the learning of the mean estimator is equivalent to regular mean-squared error. However, this approach means that learnt features may not support variance predictions well, an issue we discuss in the following section.

\subsection{Representation Issues}
\label{sec:feature_collapse}
 
In modern deep learning, we hypothesize that representations learnt for mean predictions may not support variance predictions. The first reason is due to representation collapse, whilst the other is due to the geometry of the learnt features, as also discussed in \cite{nix1994learning}.

Representation collapse refers to the phenomenon where training a neural network end-to-end leads to a compression in the deeper latent representations, such that distances on the data manifold are no longer preserved. This means that the network may primarily retain information relevant to its mean predictions, potentially compromising tasks that rely on distance information. This has recently been discussed in the context of uncertainty quantification \citep{van2020uncertainty,liu2022simple,mukhoti2022deepdeterministicuncertaintysimple, jimenez2025vecchiagaussianprocessensembles,jimenez2025probabilisticskipconnectionsdeterministic}. Some works propose to regularize the latent representations such that they remain distance \textit{aware} or \textit{preserving} with respect to distances on the data manifold; however, this requires modifying the training process.

In the regression setting, we hypothesise that representation collapse may be particularly pronounced when the variance function of the mean-variance network is not allowed to affect the learning of latent representations such as in \citet{stirn2023faithful}'s \emph{faithful} approach. Intuitively, directions on the data manifold along which the predicted mean $\mu_\theta(x)$ does not vary may be eliminated by the network during mean-only learning. In other words, the latent representation learned for predicting the mean may not span the subspace of directions relevant for predicting the variance.

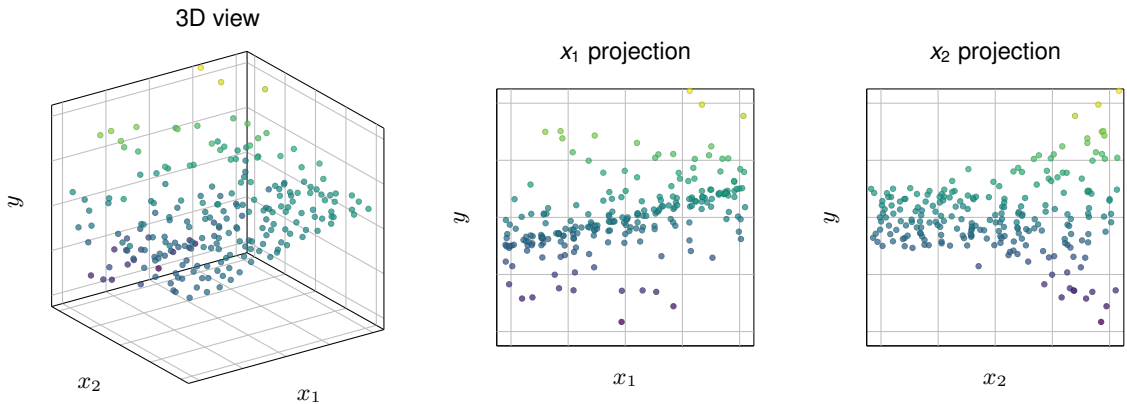
\begin{wrapfigure}[18]{R}{0.5\textwidth}
    \centering
    \resizebox{\linewidth}{!}{%
        \begin{tikzpicture}
            \newcommand{\myScatter}{
                \pgfmathsetseed{1}
                \addplot3[
                    only marks,
                    scatter,
                    mark size=0.8pt, 
                    opacity=0.75,
                    samples=15,
                    domain=-4:4,
                    domain y=0:4
                ]
                (
                {x + 0.2*rand},
                {y + 0.2*rand},
                {0.5*x + 0.5*y^2*rand}
                );
            }
            \pgfplotsset{
                my3Dstyle/.style={
                        title style={font=\sffamily\sansmath\small},
                        grid=major,
                        xtick={-4,-2,0,2,4},
                        ytick={0,1,2,3,4},
                        ztick={-8,-4,0,4,8},
                        xticklabels=\empty,
                        yticklabels=\empty,
                        zticklabels=\empty,
                        tick style={draw=none},
                        label style={font=\small},
                        xmin=-4.5, xmax=4.5,
                        ymin=-.25, ymax=4.25,
                        zmin=-9, zmax=9,
                        xlabel={$x_1$},
                        ylabel={$x_2$},
                        zlabel={$y$}
                    }
            }
            \begin{groupplot}[
                    group style={
                            group size=3 by 1,
                            horizontal sep=1.0cm, 
                            vertical sep=1cm
                        },
                    width=4cm, height=4cm,
                    colormap/viridis,
                    z buffer=sort,
                    point meta=z
                ]

                \nextgroupplot[my3Dstyle, view={-35}{25}, title={3D view}]
                \myScatter

                \nextgroupplot[my3Dstyle, view={0}{0}, title={$x_1$ projection}, ytick=\empty]
                \myScatter

                \nextgroupplot[my3Dstyle, view={90}{0}, title={$x_2$ projection}, xtick=\empty]
                \myScatter

            \end{groupplot}
        \end{tikzpicture}%
    }
    \caption{Illustrative plot of data where $x_1$ dimension fully explains $\mu(x)$, whilst $x_2$ fully explains $\sigma(x)$. (Left): 3D View of data distributed with $y=ax_1+bx_2^2\epsilon$ with $\epsilon\sim\mathcal{N}(0,1)$. (Middle): Projection of $y$ onto $x_1$ where mean increases but variance is constant. (Right): Projection of $y$ onto $x_2$ where variance of data increases, but mean remains constant.}
    \label{fig:dataexample}
\end{wrapfigure}

To make this precise, let the data $\mathcal{X}$ lie on a manifold $\mathcal{M}$, and assume there exists ground-truth target functions $\mu:\mathcal{M}\rightarrow\mathbb{R}$ and $\sigma:\mathcal{M}\rightarrow\mathbb{R}^+$ which can be exactly decomposed as
\begin{equation}
    \mu(x)=H(h(x)), \quad \sigma(x)=G(g(x)),
\end{equation}
where $h: \mathcal{M}\rightarrow \mathcal{H}$ and $g:\mathcal{M}\rightarrow \mathcal{G}$ are nonlinear embeddings, $\mathcal{H}$ and $\mathcal{G}$ are minimal latent vector spaces, and $H:\mathcal{H}\rightarrow \mathbb{R}$, $G:\mathcal{G}\rightarrow \mathbb{R}^+$ are (generalized) linear models. This corresponds to neural networks with non-linear backbones and linear output layers. If we first learn $\mu_\theta(x)=H_\theta(h_\theta(x))$ and use the latent representation $h_\theta(x)$ learned for $\mu$ to predict the variance, $\sigma$ can be exactly represented only if $\mathcal{G}\subseteq \mathcal{H}$. In Figure~\ref{fig:dataexample} we provide a simple, illustrative example of data where this is not the case.

We now explain why representation collapse is a larger problem for models that decouple location and scale, such as heteroskedastic Gaussian regression, compared with models that do not, such as binary classification.

\emph{Binary classification}:
Consider a neural network $z_\theta(x): \mathbb{R}^d\rightarrow \mathbb{R}$, trained with binary cross entropy loss and a sigmoid activation. Suppose $z_\theta(x)$ is $L$-Lipschitz,
\begin{equation}
    |z_\theta(x)-z_\theta(x')| \le L\|x-x'\|.
\end{equation}
Let $x_0$ be a point on the decision boundary, $z_\theta(x_0)=0$. For any point $x$ at distance $r=\|x-x_0\|$ from the boundary the network output satisfies
\begin{equation}
    |z_\theta(x)|\le Lr \Rightarrow 1-s \le p(y=1|x) \le s,
\end{equation}
where $s=\sigma(Lr)$ and $p(y=1|x)=\sigma(z_\theta(x))$.
This shows that the smoothness of the neural network directly bounds the classifier's confidence, forcing it to express uncertainty in between classes. The same reasoning applies to any probabilistic models where location and scale are not decoupled: Smoothness of the mean function constrains the predictive uncertainty.

\emph{Heteroskedastic Gaussian regression}:
In contrast, consider neural networks $\mu_\theta(x):\mathbb{R}^d\rightarrow\mathbb{R}$, $\sigma_\theta(x): \mathbb{R}^d\rightarrow\mathbb{R}^+$, parameterizing the mean and variance of a Gaussian. Here the predictive uncertainty is determined solely by $\sigma_\theta(x)$. Crucially, smoothness of $\mu_\theta(x)$ or $\sigma_\theta(x)$ does not constrain the uncertainty that can be expressed. Therefore, even smooth networks can experience representation collapse without a corresponding increase in predicted uncertainty, making models with decoupled location and scale particularly vulnerable.

We finally note that the requirement $\mathcal{G}\subseteq\mathcal{H}$ can also be violated when the variance-relevant information is retained in $h_\theta(x)$, but only as $g=\varphi\circ h_\theta$ for some nonlinear $\varphi$. That is, because $G(h_{\theta}(x))=\mathrm{softplus}(w^\top h_\theta(x))$ depends on $h_\theta(x)$ only through a single projection and is monotone along it, it cannot express a variance that is non-linear in $h_{\theta}$, e.g., one that grows in all directions away from a region of the latent space. Empirically, we observe that intermediate representations have a linear geometry that allows linear variance heads to achieve strong downstream performance.

\section{Method}
\label{sec:methods}
Keeping these pitfalls in mind, we propose a method for aleatoric uncertainty estimation in deep learning models for regression tasks. Specifically, we propose learning a linear variance head \textit{post hoc} on a hold-out dataset whilst keeping the mean predictor, and by extension, backbone, fixed. 
We later show that this simple linear design on intermediate representations is sufficient for obtaining the best performance in practice. A central piece to our method is to use one or more intermediate \textit{latent representations} (denoted $z^l$) as inputs to the linear variance head, rather than the last latent representation $z^L$ only \citep{kristiadi2020being,kristiadi2022posteriorrefinementimprovessample,stirn2023faithful}.

The general form for this linear variance head is
\begin{equation}
    \label{eq:linear_variance_prediction}
    \sigma^2_{\phi}(x^*)=\text{softplus}\left(\sum_{l\in L_\sigma}W^T_lz^l(x^*)\right),
\end{equation}
i.e. a sum of linear projections on the latent representations $z^l$ learnt during mean estimator training. Here $L_\sigma$ indicates the layers chosen for inputs to the variance head. The parameters $\phi = \{W^T_{l}\}_{l \in L_\sigma}$ are learnt using the NLL loss in Eq.~\eqref{eq:nll_loss} on hold-out data, with a fixed mean estimator $\mu_{\theta}(x)$ learnt on training data. Generally, we use the notation $\sigma^2_{l}(x)$ to denote the uncertainty estimator learnt with $L_{\sigma}=\{l\}$, i.e., on the $l$'th latent representation only and $\sigma^2_{L_\sigma}(x)$ to denote the estimator learnt on all latent representations, i.e., with $L_{\sigma}=\{1,2,\dots,L\}$.

We extend our method by ensembling independently learned uncertainty estimators built on individual latent representations, which we show further improves performance. Specifically, we form the predictive distribution as a Gaussian scale mixture given by
\begin{equation}
    \label{eq:variance_ensemble}
    p(y^*|x^*) = \frac{1}{|L_\sigma|} \sum_{l \in L_\sigma}  \mathcal{N}(y^*|\mu_{\theta}(x^*), \sigma^2_l(x^*)) \,,
\end{equation}
where we emphasize that all components have the same mean. We propose this modification as prior work consistently indicates that ensembles can improve performance by reducing variance \citep{batchelor1995forecaster,krogh1997statistical}, in this case at a minimal additional cost.

Whilst simple, our method addresses all of the previously identified problems of deep heteroskedastic regression jointly. Firstly, our method resolves the issue of underestimated variance outputs described in Sec.~\ref{sec:underestimated_residuals} by directly learning the uncertainty estimator on hold-out data where (benign) overfitting has not occurred. Secondly, we address optimisation issues by fitting the variance head post hoc and independently from the mean estimator. Thirdly, whilst this could introduce the representation collapse and geometry issues as described in Sec.~\ref{sec:feature_collapse}, our method mitigates this problem by utilising intermediate latent representations for the uncertainty predictor as has been done in previous works \citep{jimenez2025vecchiagaussianprocessensembles}. As a bonus, the method is highly practical, as it does not require interventions at training time nor interferes with the mean estimator, only requires a standard forward pass at prediction time with minimal additional parameter cost, and is easy to implement as it only requires writing a hook in PyTorch \citep{Ansel_PyTorch_2_Faster_2024} to extract latent representations and use a linear layer on top of these---all of which makes it especially practical in the context of large foundation models.

\section{Experiments}

\begin{table}[t]
    \centering
    \caption{Test mean absolute errors (MAE) and negative log-likelihoods (NLL) for baseline methods and our post-hoc variance ensemble. Results are averages over three seeds (five for tabular datasets) with best results highlighted in bold font. The standard error of the mean is reported in parenthesis in the unit of the last significant digit.}
    \label{tab:qm9_results}
    \resizebox{\textwidth}{!}{%
        \begin{tabular}{llllllllllllll}
\toprule
 &  & \multicolumn{6}{c}{$\textrm{MAE}\downarrow$} & \multicolumn{6}{c}{$\textrm{NLL}\downarrow$} \\
\cmidrule(lr){3-8} \cmidrule(lr){9-14}
 &  & \makecell{$\textrm{Homo.}$\\$\textrm{NLL}$} & \makecell{$\textrm{Naïve}$\\$\textrm{NLL}$} & \makecell{$\textrm{Nat.}$\\$\textrm{NLL}$} & $\beta\textrm{-NLL}$ & $\textrm{Faithful}$ & \makecell{$\textrm{Post hoc}$\\$\textrm{(ours)}$} & \makecell{$\textrm{Homo.}$\\$\textrm{NLL}$} & \makecell{$\textrm{Naïve}$\\$\textrm{NLL}$} & \makecell{$\textrm{Nat.}$\\$\textrm{NLL}$} & $\beta\textrm{-NLL}$ & $\textrm{Faithful}$ & \makecell{$\textrm{Post hoc}$\\$\textrm{(ours)}$} \\
 &  &  &  &  &  &  &  &  &  &  &  &  &  \\
\midrule
\multirow{10}{*}{\rotatebox{90}{QM9}} & $\alpha$ & 0.077 {\scriptsize(26)} & 0.044 {\scriptsize(0)} & 0.045 {\scriptsize(0)} & \textbf{0.043 {\scriptsize(0)}} & 0.050 {\scriptsize(1)} & 0.077 {\scriptsize(26)} & -0.412 {\scriptsize(143)} & -1.36 {\scriptsize(5)} & -1.29 {\scriptsize(2)} & \textbf{-1.40 {\scriptsize(1)}} & -0.665 {\scriptsize(31)} & -1.04 {\scriptsize(30)} \\
 & $\epsilon_{\textrm{HOMO}}$ & 26.6 {\scriptsize(1)} & 25.4 {\scriptsize(2)} & 25.8 {\scriptsize(2)} & \textbf{25.3 {\scriptsize(2)}} & 26.7 {\scriptsize(2)} & 26.6 {\scriptsize(1)} & -1.69 {\scriptsize(0)} & -1.94 {\scriptsize(2)} & -1.91 {\scriptsize(2)} & -1.92 {\scriptsize(2)} & -1.72 {\scriptsize(5)} & \textbf{-1.96 {\scriptsize(0)}} \\
 & $\epsilon_{\textrm{LUMO}}$ & 19.1 {\scriptsize(1)} & 19.1 {\scriptsize(1)} & 19.0 {\scriptsize(1)} & \textbf{18.9 {\scriptsize(2)}} & 19.2 {\scriptsize(1)} & 19.1 {\scriptsize(1)} & -2.00 {\scriptsize(1)} & -2.28 {\scriptsize(1)} & -2.26 {\scriptsize(1)} & -2.27 {\scriptsize(2)} & -1.85 {\scriptsize(12)} & \textbf{-2.30 {\scriptsize(1)}} \\
 & $\Delta \epsilon$ & 43.4 {\scriptsize(2)} & 43.6 {\scriptsize(4)} & 43.2 {\scriptsize(3)} & \textbf{42.8 {\scriptsize(4)}} & 43.4 {\scriptsize(1)} & 43.4 {\scriptsize(2)} & -1.29 {\scriptsize(0)} & -1.42 {\scriptsize(1)} & -1.43 {\scriptsize(1)} & -1.43 {\scriptsize(1)} & -1.34 {\scriptsize(1)} & \textbf{-1.48 {\scriptsize(0)}} \\
 & $\textrm{ZPVE}$ & 1.28 {\scriptsize(1)} & 1.25 {\scriptsize(0)} & 1.23 {\scriptsize(0)} & \textbf{1.23 {\scriptsize(0)}} & 1.28 {\scriptsize(1)} & 1.28 {\scriptsize(1)} & -4.15 {\scriptsize(5)} & -4.81 {\scriptsize(3)} & -4.92 {\scriptsize(2)} & -4.95 {\scriptsize(2)} & -4.12 {\scriptsize(4)} & \textbf{-4.97 {\scriptsize(1)}} \\
 & $U_0$ & 5.64 {\scriptsize(4)} & 5.89 {\scriptsize(6)} & 5.96 {\scriptsize(3)} & \textbf{5.48 {\scriptsize(2)}} & 5.62 {\scriptsize(4)} & 5.64 {\scriptsize(4)} & -2.85 {\scriptsize(1)} & -2.93 {\scriptsize(4)} & -2.98 {\scriptsize(4)} & -2.86 {\scriptsize(13)} & -2.32 {\scriptsize(20)} & \textbf{-3.46 {\scriptsize(2)}} \\
 & $U$ & \textbf{5.59 {\scriptsize(8)}} & 6.00 {\scriptsize(12)} & 5.86 {\scriptsize(5)} & 5.78 {\scriptsize(7)} & 5.64 {\scriptsize(5)} & 5.59 {\scriptsize(8)} & -2.84 {\scriptsize(1)} & -2.86 {\scriptsize(9)} & -2.99 {\scriptsize(4)} & -2.88 {\scriptsize(5)} & -2.60 {\scriptsize(19)} & \textbf{-3.45 {\scriptsize(1)}} \\
 & $H$ & 5.61 {\scriptsize(6)} & 6.01 {\scriptsize(12)} & 6.06 {\scriptsize(4)} & \textbf{5.56 {\scriptsize(4)}} & 5.64 {\scriptsize(7)} & 5.61 {\scriptsize(6)} & -2.85 {\scriptsize(2)} & -2.85 {\scriptsize(17)} & -2.94 {\scriptsize(11)} & -2.93 {\scriptsize(4)} & -1.92 {\scriptsize(44)} & \textbf{-3.46 {\scriptsize(2)}} \\
 & $G$ & 7.21 {\scriptsize(6)} & 7.55 {\scriptsize(6)} & 7.47 {\scriptsize(16)} & \textbf{7.11 {\scriptsize(9)}} & 7.21 {\scriptsize(2)} & 7.21 {\scriptsize(6)} & -2.74 {\scriptsize(2)} & -2.84 {\scriptsize(0)} & -2.85 {\scriptsize(7)} & -2.90 {\scriptsize(4)} & -1.48 {\scriptsize(61)} & \textbf{-3.16 {\scriptsize(2)}} \\
 & $c_{\textrm{v}}$ & 0.024 {\scriptsize(0)} & 0.024 {\scriptsize(0)} & 0.024 {\scriptsize(0)} & \textbf{0.024 {\scriptsize(0)}} & 0.024 {\scriptsize(0)} & 0.024 {\scriptsize(0)} & -1.91 {\scriptsize(0)} & -2.02 {\scriptsize(0)} & -1.98 {\scriptsize(0)} & -2.03 {\scriptsize(1)} & -1.91 {\scriptsize(1)} & \textbf{-2.04 {\scriptsize(1)}} \\
\midrule
\multicolumn{2}{l}{UTKFace} & 4.88 {\scriptsize(1)} & \textbf{4.86 {\scriptsize(2)}} & 4.90 {\scriptsize(2)} & 4.87 {\scriptsize(2)} & 4.88 {\scriptsize(1)} & 4.88 {\scriptsize(1)} & 3.43 {\scriptsize(1)} & 3.37 {\scriptsize(5)} & 3.41 {\scriptsize(4)} & 3.32 {\scriptsize(4)} & 3.60 {\scriptsize(10)} & \textbf{3.25 {\scriptsize(0)}} \\
\midrule
\multicolumn{2}{l}{Cal. Housing} & 0.308 {\scriptsize(3)} & 0.291 {\scriptsize(2)} & \textbf{0.289 {\scriptsize(1)}} & 0.304 {\scriptsize(2)} & 0.303 {\scriptsize(2)} & 0.308 {\scriptsize(3)} & 0.663 {\scriptsize(7)} & 0.546 {\scriptsize(34)} & 0.596 {\scriptsize(43)} & \textbf{0.434 {\scriptsize(9)}} & 0.494 {\scriptsize(6)} & 0.519 {\scriptsize(13)} \\
\multicolumn{2}{l}{Year} & 5.99 {\scriptsize(2)} & \textbf{5.92 {\scriptsize(2)}} & 5.93 {\scriptsize(0)} & 5.94 {\scriptsize(1)} & 5.99 {\scriptsize(1)} & 5.99 {\scriptsize(2)} & 3.60 {\scriptsize(0)} & 3.38 {\scriptsize(1)} & 3.38 {\scriptsize(1)} & \textbf{3.35 {\scriptsize(0)}} & 3.38 {\scriptsize(0)} & 3.36 {\scriptsize(0)} \\
\bottomrule
\end{tabular}

    }
\end{table}

We conduct a series of experiments on tabular, image, and molecular datasets, to showcase that our proposed method performs on par with or better than end-to-end trained mean-variance networks, despite its simplicity. We then conduct a number of ablation studies to showcase that our variance estimator design generally cannot be beaten by more complex variance predictors and is simple and robust. Finally, we show that these results and ablations generalize to large-scale machine learning interatomic potentials trained on OMol25 \citep{levine2025openmolecules2025omol25}.

\paragraph{Datasets and Models}
We compare against baselines by using the PaiNN graph neural network \citep{schutt2021equivariant} for molecular property prediction on the QM9 dataset \citep{moleculenet}, a ResNet model \citep{he2016deep} for age regression on the UTKFace image dataset \citep{zhifei2017cvpr}, and the FT-Transformer model \citep{gorishniy2021revisiting} on the tabular datasets California Housing \citep{pace1997sparse} and Year \citep{year_prediction_msd_203}. In the pretrained model experiments, we experiment on the OMol25 dataset using FAIR's UMA model \citep{UMA} and the AllScAIP model \citep{qu2026recipescalableattentionbasedmlips}. Both the UMA model and the AllScAIP model are trained on 100 million molecules from the OMol25 dataset \citep{levine2025openmolecules2025omol25} while UMA is additionally trained on OMat24 \citep{barroso2024open}, OC20 \citep{chanussot2021open}, ODAC25 \citep{sriram2025open}, and OMC25 \citep{gharakhanyan2025open}, rendering end-to-end mean-variance network training infeasible.

\paragraph{Baselines \& Experimental Setup}
In the comparison experiments, we compare our method against homoskedastic regression and heteroskedastic regression using a naïve Gaussian negative log-likelihood (Naïve NLL), a Gaussian NLL using the natural parameterisation (Natural NLL) \citep{immer2024effective}, $\beta$-NLL \citep{seitzer2022pitfalls}, and Faithful heteroskedastic regression (Faithful) \citep{stirn2023faithful}. All of these methods learn a mean and variance function jointly on the training data, while ours only learn a variance function on a hold-out dataset. To enable a fair comparison, we fit a constant scaling factor for the baselines' variance functions on the hold-out data to improve calibration as proposed by \citet{levi2020evaluating}---we again note, that without this, the NLLs of the end-to-end trained models are extremely poor (App.~\ref{app:uncalibrated_baselines}). We additionally tune the $\beta$ parameter for $\beta$-NLL by cross-validation using the hold-out data.

As our goal is to obtain useful uncertainty estimates while retaining the predictive performance of the mean, we early-stop training based on the performance of the mean on the hold-out data for all methods. Our post-hoc method is learned on top of the homoskedastic model using the ensemble approach in Eq.~\eqref{eq:variance_ensemble} with the hold-out data. The individual variance functions $\sigma^2_l(x)$ in the ensemble are learnt independently for all latent representations of the homoskedastic network architectures, i.e., we use $L_{\sigma}=\{0,1,\dots,L\}$ in Eq.~\eqref{eq:variance_ensemble}, where $z^L$ is the last-layer representation.

In the OMol25 experiments, training from scratch with the baseline methods is infeasible, and we are therefore limited to a comparison against homoskedastic regression, where the variance scale is again learned on hold-out data. The OMol25 experiments are conducted to showcase that the results and observations from the comparison experiments generalises to larger models and dataset and that we can equip large-scale models with useful uncertainty estimates in a practical way without significant computational overhead. For additional training details, we refer to App.~\ref{sec:training_details}.

\subsection{Baseline Comparison Experiments}
\label{sec:qm9_experiments}

\begin{figure}[t]
    \subcaptionbox{QM9\label{fig:qm9_latent}}
    {\includegraphics[width=0.32\linewidth]{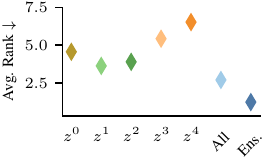}}
    \subcaptionbox{UTKFace\label{fig:utkface_latent}}
    {\includegraphics[width=0.32\linewidth]{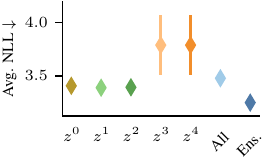}}
    \subcaptionbox{UMA on OMol25\label{fig:uma_latent}}
    {\includegraphics[width=0.32\linewidth]{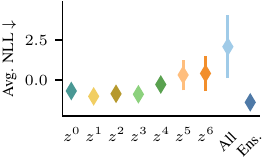}}
    \caption{Test NLL for fitting the uncertainty estimator using different individual representations $z^{l}$, all representations (All), or an ensemble of the individual estimators (Ens.). For the QM9 dataset, the NLL-based average rank across targets is showcased. The rankings are computed for every combination of target and seed and averaged. Error bars indicate one standard error of the mean.}
\label{fig:latent_ablation}
\end{figure}

We firstly present results from the experiments comparing our method to existing methods for end-to-end training of mean-variance networks. In \cref{tab:qm9_results}, we see that all methods achieve similar MAE on the test data with $\beta$-NLL achieving slightly better means on most targets. When comparing NLLs, we note that across all targets and datasets except for the Year dataset, our post-hoc method is at least on-par with existing end-to-end methods, but also often \textit{superior}. To summarize, our post-hoc variance ensemble retains predictive performance (in terms of the MAE) while offering at least competitive, but often better, NLLs. In App.~\ref{app:additional_results_qm9} we provide additional results and analysis of the models trained on QM9, including OOD evaluations where our method similarly is on-par with end-to-end trained methods, and calibration plots that show that our method learns input-varying uncertainties on targets where the end-to-end trained models fail to do so. In App.~\ref{app:isotonic}, we also recalibrate the baseline methods using isotonic regression on the hold-out data instead of a simple variance rescaling, but note that isotonic regression does not significantly improve the baselines. In App.~\ref{app:conformal}, we further compare all methods to split and locally adaptive conformal prediction in terms of coverage, interval width and interval score, where we find that our post-hoc variance estimators provide a favourable trade-off between coverage and width.

\subsection{Ablation Studies}
Having now shown it is possible to learn good aleatoric uncertainty estimation post-hoc using our proposed method, we now investigate and provide evidence that the design of our variance estimator is simple yet robust, and provide empirical evidence that the pitfalls presented in \cref{sec:pitfalls} must be addressed for success in post-hoc deep heteroskedastic regression. We have already illustrated that it empirically is necessary to use hold-out data to properly learn variance outputs, evidenced by end-to-end trained baselines having extremely poor NLLs without temperature scaling (Appendix \ref{app:uncalibrated_baselines}).
In the following we show that: (1) The choice of latent representation is important and that intermediate latent representations generally outperform early or last-layer representations, (2) Our ensembling methods outperforms other end-to-end trained baselines even at very small hold-out dataset sizes, (3) Our method is surprisingly robust to the only hyperparameter choice in the linear layer, i.e.\ the weight decay, and finally (4) that introducing additional complexity in the post-hoc learnt variance predictor generally does not provide improvement over the simple linear variance predictor.

\begin{figure}[t]
    \subcaptionbox{QM9 $U$ target \label{fig:qm9_learning_curve}}
    {\includegraphics[width=0.45\linewidth]{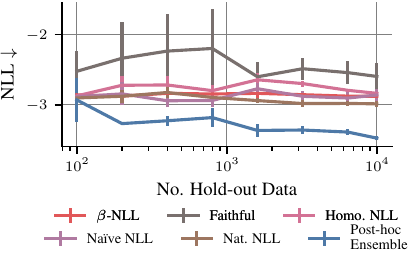}}
    \subcaptionbox{UMA on OMol25\label{fig:omol_learning_curve}}
    {\includegraphics[width=0.45\linewidth]{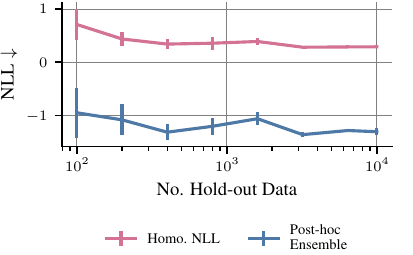}}
    \caption{Effect of varying hold-out dataset size. Results are test NLL averaged over three seeds with error bars indicating one standard error of the mean.}
    \label{fig:learning_curve}
\end{figure}

\paragraph{Choice of Latent Representation}
\label{sec:latent_representation}
We firstly investigate how the choice of the latent representation $z^{l}$ used in the variance predictor affects performance.
In \cref{fig:latent_ablation}, we vary the representation $z^l$ used in the linear variance predictor $\sigma^2_{l}(x)$  across the depth of the network, where higher superscripts denote later layers. We also compare against the uncertainty estimator $\sigma^2_{L_{\sigma}}(x)$ learnt on the concatenation of all latent representations (referred to as "All" in \cref{fig:latent_ablation}), i.e.\ $L_{\sigma}=\{0, 1, \dots, L\}$, and the ensemble version that we presented in the comparison experiments.

Across all panels in \cref{fig:latent_ablation}, we notice the same pattern: intermediate latent layers provide better NLL compared to last-layer latent representations which often fail to generalize well. However, the ensemble of individual predictors generally performs the best, significantly outperforming the other estimators. We have thus shown that simply relying on last-layer latent representations is not sufficient when performing post-hoc variance estimation. For completeness, we also ablate using learnt latent representations by instead using random Fourier features as input to the linear variance predictors. This performs worse than using the learnt features and is also sensitive to the lengthscale choice which would require tuning---see Table \ref{tab:rff_features} in Appendix \ref{app:additional_results_qm9}.

\begin{wrapfigure}[21]{R}{0.5\textwidth}
    \includegraphics[width=0.5\textwidth]{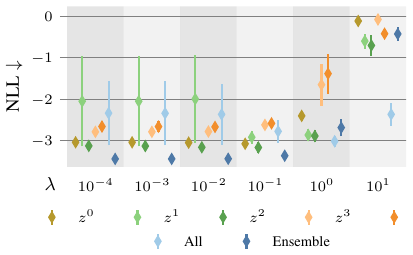}
    \caption{The effect of varying the weight decay parameter $\lambda$ when fitting the linear uncertainty estimator using different individual representations $z^{l}$, all representations, or an ensemble of the individual estimators. Results are test NLL averages over three seeds on the QM9 $U$ target and are plotted with one standard error.}
    \label{fig:weight_decay_ablation}
\end{wrapfigure}

\paragraph{Hold-Out Dataset Size \& Weight Decay Sensitivity}
\label{sec:hold_out_size}

Our method naturally depends on the hold-out data used for fitting the variance function, and one might hypothesise that less hold-out data is required to recalibrate end-to-end trained networks with temperature scaling. We therefore examine the effect of varying the size of the hold-out data and compare to the baselines. Naturally, the optimal regularization strength depends on the size of the hold-out dataset. Consequently, we do not use weight decay in this setting, but instead employ an early-stopping strategy, which corresponds to using optimised weight decay \citep{bishop1995neural}. 
For details on this approach see App.~\ref{sec:training_details}. In \cref{fig:qm9_learning_curve}, we see the test NLL on the QM9 $U$ target as a function of the hold-out dataset size. Here, we observe that the baselines are relatively invariant to the hold-out dataset size. However, very surprisingly we find that our post-hoc ensemble outperforms the end-to-end trained networks even for 200 hold-out datapoints which is $\approx0.2\%$ of the training data. In \cref{fig:weight_decay_ablation} we provide evidence that our method is very robust to the only hyperparameter choice in the linear variance predictor, i.e. the weight decay. We note that most representations are largely insensitive to $\lambda$ with the ensembling version showing the highest robustness, although a too large $\lambda$ is detrimental for all representations.

\begin{wrapfigure}[17]{R}{0.5\textwidth}
    \centering
    \includegraphics[width=0.5\textwidth]{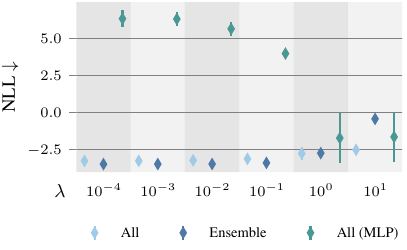}
    \caption{Ablation of variance head on the QM9 target $U_0$. Error bars indicate one standard error of the mean computed over seeds.}
    \label{fig:variance_head_ablation}
\end{wrapfigure}

\paragraph{Variance Head Ablation}
A logical question is whether adding additional capacity in the variance head will lead to increased performance. We investigate this by changing the variance head of the "All" variance estimator to be a 1-layer MLP, and evaluate performance on the QM9 $U_0$ target. In \cref{fig:variance_head_ablation}, we show that even when tuning for weight decay, the "All" (MLP) version performs worse than both the ensembled linear variance models and the "All" linear variance model. In App.~\ref{app:last_layer}, we find that, across all 13 targets, an MLP on the last-layer representation can close the gap to the best linear intermediate head, but only for a well-chosen weight decay, and fails badly otherwise. We thus argue that low complexity of the variance model is an equally performant but far simpler variance predictor, which makes it much more practical to use.

\subsection{OMol25}
\label{sec:omol_experiments}
Finally, we investigate if our observations so far generalise to the large-scale UMA and AllScAIP models. We use three random (but disjoint) subsets of 20,000 datapoints from the OMol25 validation dataset and split each of them into equally sized hold-out and test datasets and report results as averages over the three subsets. In \cref{tab:omol_results}, we report the test MAE, NLL, and ECE for energy prediction with both UMA and AllScAIP using homoskedastic regression, our post-hoc linear variance predictor using all latent representations, and our post-hoc variance ensemble. In \cref{fig:uma_latent}, we show results from testing whether intermediate latent representations provide better generalization for the UMA model, and in \cref{fig:omol_learning_curve}, we show the results from the hold-out dataset size sensitivity experiment using UMA on Omol25. Even here the same patterns generally repeat. Our post-hoc method with ensembled linear variance predictor outperforms the "All" version for both UMA and AllScAIP in terms of NLL, although the ECE of the "All" version is better.
\begin{wraptable}[20]{R}{0.5\textwidth}
    \centering
    \caption{Test metrics on the OMol25 dataset for homoskedastic regression, the post-hoc linear variance predictor using all latent representations, and the post-hoc variance ensemble fitted on top of the trained UMA and AllScAIP models. Results are averages over three seeds and subsets of the full data with best results highlighted in bold font. The standard error of the mean is reported in parenthesis in the unit of the last significant digit.}
    \label{tab:omol_results}
    \resizebox{.4\textwidth}{!}{%
    \begin{tabular}{llcc}
\toprule
\multicolumn{2}{l}{} & \multicolumn{2}{c}{$\textrm{Model}$} \\
\cmidrule(lr){3-4}
\multicolumn{2}{l}{$\textrm{Metric}$} & $\textrm{UMA}$ & $\textrm{AllScAIP}$ \\
\midrule
\multicolumn{2}{l}{$\textrm{MAE}\downarrow$} & 60.7 {\scriptsize(9)} & 42.2 {\scriptsize(1)} \\
\cmidrule(lr){1-4}
\multirow{3}{*}{$\textrm{NLL}\downarrow$} & $\textrm{Homo. NLL}$ & 0.290 {\scriptsize(20)} & -0.354 {\scriptsize(52)} \\
 & $\textrm{All}$ & 2.07 {\scriptsize(193)} & -1.59 {\scriptsize(23)} \\
 & $\textrm{Ensemble}$ & \textbf{-1.36 {\scriptsize(4)}} & \textbf{-1.94 {\scriptsize(4)}} \\
\cmidrule(lr){1-4}
\multirow{3}{*}{$\textrm{ECE}\downarrow$} & $\textrm{Homo. NLL}$ & 0.209 {\scriptsize(5)} & 0.186 {\scriptsize(9)} \\
 & $\textrm{All}$ & \textbf{0.105 {\scriptsize(17)}} & \textbf{0.025 {\scriptsize(1)}} \\
 & $\textrm{Ensemble}$ & 0.133 {\scriptsize(8)} & 0.053 {\scriptsize(1)} \\
\bottomrule
\end{tabular}

    }
\end{wraptable}

In the ablation studies, we firstly see that for the UMA model, intermediate latent representations generalize better individually, but still gets beaten by the ensembled version. Secondly, we again note that even at very small hold-out sizes, our ensembled version is surprisingly robust and performs significantly better than the homoskedastic baseline. We provide additional analysis and experimental results in Appendix \ref{app:additional_results_omol25}.

\begin{wrapfigure}[11]{R}{.5\textwidth}
    \vspace{-2ex}
    \begin{minipage}{.5\linewidth}
        \centering
        \includegraphics[width=\linewidth]{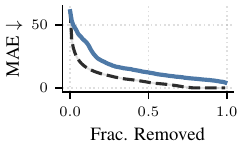}
    \end{minipage}%
    \begin{minipage}{.5\linewidth}
        \centering
        \includegraphics[width=\linewidth]{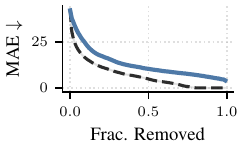}
    \end{minipage}
    \caption{Calibration plots: Avg. MAE after removing samples ranked by predicted vs. true error. (left): UMA. (right): AllScAIP.}
    \label{fig:omol_calibration}
\end{wrapfigure}

Finally, we qualitatively investigate the calibration of the post-hoc variance ensemble fitted on top of UMA and AllScAIP. In \cref{fig:omol_calibration}, we observe that the selective prediction curve for the post-hoc variance ensemble on both UMA and AllScAIP follows the oracle remarkably well\footnote{For details on how these plots are generated, see \cref{fig:app_calibration_plots_methods}.}. The predictive uncertainty (rank) correlates extremely well with the prediction error on the test data. Whilst not a guarantee for success, such uncertainties are required for effective sequential experimental design.

\section{Discussion}
\label{sec:discussion}
\paragraph{Performance} Throughout our baseline comparison experiments, we observed that our ensembled post-hoc variance estimators performed at minimum on par with end-to-end trained mean-variance networks, but in many cases outperformed them in terms of NLL. In the ablation studies, we showed that our proposed method is highly robust to different weight decay values and requires the same amount of hold-out data that end-to-end trained methods do. We showed that our design choices, e.g. using intermediate latent representations and only linear layers for variance prediction performs similarly to using tuned MLPs as variance predictors. We then verified that these results also extend to large pretrained machine learning interatomic potentials on OMol25 using the UMA and AllScAIP models. Based on this evidence, we argue that our proposed method for variance estimation in deep heteroskedastic regression is an equally viable (if not better), route to efficient aleatoric uncertainty estimation using neural networks, in particular in the era of foundation models.

\paragraph{(Lack of) Optimisation Issues} An unexpected observation from our experiments, is that existing end-to-end mean-variance training methods appear to not suffer from previously described optimisation issues. Although the end-to-end trained models often shrink their variance predictions to a near constant $0$, we observed that they often preserve the correct rankings of the variances (see \cref{fig:app_calibration_plots_methods} in App.~\ref{app:additional_results_qm9}) although on a wrong scale, but that the correct scale can simply be recovered by temperature scaling. We hypothesise that a common pitfall when training mean-variance networks may be to early-stop based on, or optimise for, validation NLL, leading to poor mean estimates which gets observed as optimization issues.

\paragraph{Relation to Previous Works}
Our method is closest to phase II of \citet{nix1994learning}, which fits a variance network with the Gaussian NLL on held-out data on top of a frozen, shallow mean network. We differ in that we (i) use all intermediate representations of a deep network, (ii) deliberately restrict the variance predictors to linear heads, (iii) ensemble these across layers, and (iv) omit their final phase of joint training, which makes our method purely post hoc and applicable to pretrained foundation models. Our method also relates to \citet{jimenez2025vecchiagaussianprocessensembles}, but introduces no additional hyperparameters and essentially no inference-time or memory cost. We elaborate more on these differences in App.~\ref{app:diff_vecchia}.

\paragraph{Limitations }

Whilst we generally find our method to work well, we did observe occasional experimental results that were different from our general observations. On the California Housing tabular dataset experiments, our method was outperformed by the $\beta$-NLL method, and in the AllScAIP latent representaion ablation (App.~\ref{app:additional_results_omol25}), we found all latent representations to provide equally good test NLL. Additionally, we also occasionally find that the "All" variation outperforms the Ensemble variation, especially in terms of out-of-distribution and ECE metrics.
This indicates a trade-off between sharper variance predictions achieved by the "All" version and the more robust Gaussian mixture predictions by the Ensemble variation, which is further corroborated by the coverage and interval width metrics in App.~\ref{app:conformal}.

\section{Conclusion}
In this work, we have identified and analysed the core problems related to training deep learning models for heteroskedastic regression in the overparameterised regime and devised a highly practical method that addresses these problems. We showed that our method is at minimum on-par, but often outperforms end-to-end trained mean-variance networks on multiple data modalities in terms of uncertainty quantification and that the method extends to large scale machine learning interatomic potentials trained on OMol25. We argue this work and method paves the way for applications of such large-scale models where good predictive uncertainty is required and can have immense impact.

\begin{ack} M. Jordahn, J. V. Jensen and M. N. Schmidt acknowledge support from the Novo Nordisk Foundation under grant no NNF22OC0076658 (Bayesian neural networks for molecular discovery). 
\end{ack}

\bibliography{references}
\bibliographystyle{plainnat}


\appendix
\etocdepthtag.toc{mtappendix}
\etocsettagdepth{mtmain}{none}
\etocsettagdepth{mtappendix}{subsection}
\etocstandardlines                         
\etocsettocstyle{\section*{Appendix Contents}}{}
\newpage
\tableofcontents
\newpage
\section{Training \& Experimental Details}
\label{sec:training_details}
In this section, we provide details on the experimental setups and training for the experiments in Sec.~\ref{sec:qm9_experiments} and \ref{sec:omol_experiments}. Code is available at \href{https://github.com/jonasvj/PracticalDeepHeteroskedasticRegression}{github.com/jonasvj/PracticalDeepHeteroskedasticRegression}.

\subsection{QM9}
For training PaiNN on QM9, we generally follow the procedure of \cite{schutt2021equivariant} with a few slight modifications. The QM9 dataset is split into 110,000 training points, 10,000 hold-out points, and 10,381 test points.

\paragraph{Homoskedastic Regression} We train the homoskedastic regression model using the MSE loss function for 1,000 epochs with a batch size of 100 molecules using the MAE on the hold-out data as an early-stopping criterion. We use the AdamW optimizer \cite{loshchilov2017decoupled}, a weight-decay factor $\lambda=0.01$, an initial learning rate of $5\mathrm{e}{-4}$ and decrease the learning rate with cosine annealing. We repeat training for three different random initilisations (seeds) for all targets.

\paragraph{Heteroskedastic Regression}
We follow the same procedure as for the homoskedastic model but use either the näive Gaussian negative log-likelihood, a Gaussian NLL using the natural parameterisation \cite{immer2024effective}, $\beta$-NLL \cite{seitzer2022pitfalls}, and Faithful heteroskedastic regression \cite{stirn2023faithful} as the loss function. All methods use the same backbone and the variance head is attached to the last-layer latent representation. For $\beta$-NLL, we do a grid search for $\beta$ for each seed and target with the grid $\{0.25, 0.5, 0.75, 1\}$ and use the NLL on the hold-out data as a selection criterion (after recalibrating with a constant variance scale factor).

\paragraph{Post-Hoc Calibration}
After fitting the homoskedastic models and the heteroskedastic models, we rescale their variance functions with a constant scale factor $\gamma >0$ such that the predictive distribution becomes
\begin{equation}
    p(y^*|x^*) = \mathcal{N}(y^*|\mu_\theta(x^*), \gamma \sigma_{\theta}^2(x^*)) \,.
\end{equation}
We fit $\gamma$ using the NLL criterion on the hold-out data with the L-BFGS optimizer and a learning rate of $0.01$, a maximum of 100 iterations per step, a history size of 1,000 (for estimating the Hessian), line search, and 1,000 epochs with full batches.

\paragraph{Post-Hoc Linear Variance Estimators}
The post-hoc linear variance estimators in \cref{eq:linear_variance_prediction} are fitted using the AdamW optimizer for 500 epochs on the hold-out data using an initial learning rate of $5\mathrm{e}{-4}$ and cosine annealing. The weight-decay factor $\lambda$ is set to $0.01$ except in the weight-decay sensitivity experiment where we vary it in the grid $\{1\mathrm{e}{-4}, 1\mathrm{e}{-3}, 1\mathrm{e}{-2}, 1\mathrm{e}{-1}, 1\mathrm{e}{0}, 1\mathrm{e}{1}\}$ and the hold-out dataset sensitivity experiment where we set it 0 and use early-stopping instead.

\paragraph{Post-Hoc Ensemble of Linear Variance Estimators}
We do not do any additional training or tuning for the post-hoc variance ensemble as we simply combine the variance estimators fitted using the above approach in a Gaussian mixture model using \cref{eq:variance_ensemble}.

\paragraph{Per Atom Variance Contributions} In the PaiNN architecture, molecular properties are predicted as a sum of atomic contributions, i.e.,
\begin{equation}
    \mu_{\theta}(x) = \sum_{i=1}^{N_x} W_{\mu}^T z_i^L(x) \,,
\end{equation}
where $N_x$ is the number of atoms in the input molecule $x$, $z_i^L(x) \in \mathbb{R}^{D \times 1}$ is the last layer representation of the $i$th atom in $x$, and $W_{\mu} \in \mathbb{R}^{D \times 1}$. Similarly, we predict the variance as a sum of atomic contributions in the heteroskedastic models
\begin{equation}
    \sigma^2_{\theta}(x) = \sum_{i=1}^{N_x} \mathrm{sp}(W_{\sigma}^T z_i^L(x)) \,,
\end{equation}
where $\mathrm{sp}$ is the softplus function and $W_{\sigma}$ has the same shape as $W_{\mu}$. For our post-hoc linear variance estimators, this equation becomes
\begin{equation}
    \label{eq:graph_var}
    \sigma^2_{\phi}(x) = \sum_{i=1}^{N_x} \mathrm{sp} \left(\sum_{l \in L_{\sigma}}W^T_{l} z_i^l(x) \right) \,,
\end{equation}
where $z_i^l(x)$ is $l$th latent representation of atom $i$ in molecule $x$.

\paragraph{Latent Representations}
For the PaiNN architecture, we extract five different latent representations for the different linear variance estimators. We extract the scalar atom representations after each block of message passing and updating as well as the scalar atom representations in the readout MLP. PaiNN has three blocks of message passing and updating, resulting in the representations $z^{\{0,1,2\}}$, and a two-layer readout MLP, resulting in the representations $z^{\{3,4\}}$, where $z^{4}$ is the last-layer representation. We fit variance estimators $\sigma_l^2(x)$ on each of these individual latent representations, a variance estimator $\sigma_{L_{\sigma}}^2(x)$ using all the latent representations ($L_{\sigma}=\{0,1,\dots,4\}$), and combine the individual estimators using \cref{eq:variance_ensemble} in an ensemble.

\paragraph{Out-of-Distribution Experiment}
In the OOD experiments, we use 10,000 datapoints from the test split of the Alchemy dataset \cite{chen2019alchemy} as OOD datapoints. We then use the learned variance functions $\gamma\sigma_{\theta}^2(x)$ (the baselines), $\sigma_l(x)^2$, and $\sigma_{L_{\sigma}}^2(x)$ as classifiers to discriminate between ID data (the QM9 test split) and OOD data, and evaluate them using the false positive rate at a 95\% true positive rate (FPR95) and the area under receiver operating characteristic curve (AUROC). For the Gaussian mixture model (the ensemble), we have that the variance is
\begin{equation}
    \label{eq:ensemble_variance}
    \mathbb{V}\left[\frac{1}{|L_\sigma|} \sum_{l \in L_\sigma}  \mathcal{N}\left(\mu_{\theta}(x^*), \sigma_l^2(x^*)\right) \right]  = \frac{1}{|L_\sigma|} \sum_{l \in L_\sigma} \sigma_l^2(x^*) \,,
\end{equation}
due to the component means being equal. We therefore use \cref{eq:ensemble_variance} as the OOD classifier for the post-hoc variance ensemble.

\paragraph{Hold-Out Dataset Size Experiment}
In the experiment investigating the effect of the size of the hold-out dataset, we fit the $\gamma$ parameter for the baselines as described above but only use the given hold-out subset. For our post-hoc linear variance estimators, we set the weight-decay factor $\lambda$ to 0 and instead use early-stopping to regularize the models. We do this by further splitting the given hold-out subset into a set used for fitting and one for monitoring the NLL to determine the optimal number of optimization steps, i.e., when to early-stop. When we have found the optimal number of optimization steps, we refit the linear variance estimator using the complete hold-out subset with the found optimal number of steps. We use an 80/20 ratio for this splitting. We use the same optimizer settings as described above but set the maximum number of epochs to 5,000 and use a patience of 50 epochs for the early-stopping.

\paragraph{Variance Head Complexity Experiment} To test if performance could be improved by increasing the complexity of the variance head, we trained a 1-layer MLP (with SiLU activations) on top of the "All" representation (the concatenation of all representations) to predict the variance on the QM9 targets $U_0$ and $U$. The heads are fitted with AdamW for 500 epochs with a learning rate of $5\mathrm{e}{-4}$ with cosine annealing on the hold-out data and for different weight decays $\lambda$ (the same grid as in the weight-decay experiment).

\paragraph{Random Fourier Features Experiment}
To see if the learnt representations actually contain any useful information for the variance prediction, we try to do variance prediction based on random fourier features instead of the learnt representations for the QM9 target $U_0$. As we cannot apply random Fourier features directly to the graph input $x$, we generate representations by passing the input through a randomly initialized neural network (PaiNN), extracting the representations (either the initial embeddings $z^{\text{emb}}$, the representations after the message-passing blocks $z^{\{0,1,2\}}$, or the representations in the readout network $z^{\{3,4\}}$), and computing random fourier features from these with an RBF-kernel. We repeat the experiment for a grid of the RBF length scale $\sigma^2$ and additionally repeat the experiment using raw random features without the random Fourier features extension ("No RFF").

\subsection{OMol25}

\paragraph{Data} The test split of the OMol25 dataset do not have publicly available labels. Therefore, we use the validation split to create our hold-out and test datasets. We take 10,000 random samples as our hold-out dataset and another 10,000 random samples as our test data to keep computational resources needed for testing manageable. We repeat this process three times, while ensuring there is no overlap between our samples, to get three datasets each with 10,000 hold-out datapoints and 10,000 test datapoints. We report our results as averages over these three datasets. We use the DFT total energy as our target.

\paragraph{Homoskedastic regression} For the homoskedastic model, we use the pretrained UMA and AllScAIP models as the mean estimator $\mu_{\theta}(x)$ and simply fit the variance $\gamma$ using the hold-out data using the same approach and parameters as when doing post-hoc calibration in the QM9 setup.

\paragraph{Post-Hoc Linear Variance Estimators} We fit the post-hoc linear variance estimators using the same procedure and parameters as for QM9, with the small addition that we first fit an initial variance scale $\gamma$ using the same approach as for the homoskedastic model. We do this to help optimization, i.e., not to have variance estimates that are orders of magnitude off at the beginning of optimization. This step is not needed in the QM9 setup as an initial scale is already embedded into the model.

\paragraph{Per Atom Variance Contributions} The total DFT energy is also predicted as sums of atomic contributions in the UMA and AllScAIP models, and we therefore use \cref{eq:graph_var} again to predict variances.

\paragraph{Latent Representations} For the UMA model, we extract seven different latent representations. The UMA model is based on the equivariant graph neural network eSEN \citep{fu2025learning}. The smaller variant of UMA that we use has 4 blocks of edgewise and nodewise processing, and we extract the atom scalar representations after each of the blocks, resulting in $z^{\{0,\dots,3 \}}$. The readout MLP of UMA has three layers and we extract the latent representations from this, resulting in the final three representations $z^{\{4,5,6 \}}$. We use all seven representations in the variance estimator $\sigma_{L_{\sigma}}^2(x)$ and in the ensemble model.

The AllScAIP model is an extension of the EScAIP model \citep{qu2024importance} and is an attention based architecture. We extract eight different latent representations from the model, where the first six representations $z^{\{0,\dots,5\}}$ are the atom embeddings efter each transformer block and the last 2 representations $z^{\{6,7\}}$ are the atom representations from the two-layer readout MLP.  We use all eights representations in the variance estimator $\sigma_{L_{\sigma}}^2(x)$ and in the ensemble model.

\paragraph{Hold-Out Dataset Size Experiment} In the hold-out dataset size sensitivity experiment for the UMA model, we use the exact same setup and settings as in the QM9 case (except that we also here fit an intial variance scale before optimizing the linear models).

\subsection{UTKFace}
For training ResNet50 on the UTKFace dataset for age regression, we generally follow the protocol in \cite{gustafsson2020energy}. The UTKFace dataset is split into 19,284 training points, 2,410 hold-out points, and 2,412 test points and the ResNet50 model is pretrained on ImageNet.

\paragraph{Homoskedastic Regression} We train the homoskedastic regression model using the MSE loss function for 200 epochs with a batch size of 32 using the MAE on the hold-out data as an early-stopping criterion. We use the AdamW optimizer \citep{loshchilov2017decoupled}, a weight-decay factor $\lambda=0.001$, an initial learning rate of $1\mathrm{e}{-4}$ and decrease the learning rate with cosine annealing. We repeat training for three different random initilisations (seeds).

\paragraph{Heteroskedastic Regression}
Like in the QM9 experiments, we use the same settings as for the homoskedastic model but use the loss functions and/or training modifications from the baseline methods, and tune $\beta$ for $\beta$-NLL using the same grid as in the QM9 experiments. 

\paragraph{Post-Hoc Calibration}
After fitting the homoskedastic models and the heteroskedastic models, we rescale their variance functions with a constant scale factor $\gamma >0$ using the same approach and settings as in the QM9 experiments.

\paragraph{Post-Hoc Linear Variance Estimators}
The post-hoc linear variance estimators in \cref{eq:linear_variance_prediction} are fitted using the AdamW optimizer for 200 epochs on the hold-out data using the same batch size, learning rate, and weight decay as the end-to-end trained models.

\paragraph{Post-Hoc Ensemble of Linear Variance Estimators}
Like in the QM9 experiments, we do not do any additional training or tuning for the post-hoc variance ensemble as we simply combine the variance estimators fitted using the above approach in a Gaussian mixture model using \cref{eq:variance_ensemble}.

\paragraph{Latent Representations}
For the ResNet50 architecture, we extract five different latent representations for the different linear variance estimators. We extract the feature maps after each of the four bottleneck blocks $z^{\{0,...,3\}}$ and the last layer representation $z^{4}$. The feature maps $z^{\{0,...,3\}}$ have dimension $C \times H \times W$. To reduce the dimensionality of these, we average-pool the spatial dimensions before feeding them to the linear variance predictors.

\subsection{Tabular Datasets (California Housing and Year)}

For training the FT-Transformer on the tabular datasets California Housing ($D_{\text{input}}=8$, $N_{\text{train}}=13,209$, $N_{\text{hold-out}}=3,303$, $N_{\text{test}}=4,128$) and Year ($D_{\text{input}}=90$, $N_{\text{train}}=370,972$, $N_{\text{hold-out}}=92,743$, $N_{\text{test}}=51,630$), we use the default model settings (non-optimized) and generally follow the protocol in \cite{gorishniy2021revisiting}. 

\paragraph{Homoskedastic Regression} We train the homoskedastic regression model using the MSE loss function for 10,000 epochs using the RMSE on the hold-out data as an early-stopping criterion with a patience of 17 epochs. We use the AdamW optimizer with a weight-decay factor $\lambda=1\mathrm{e}{-5}$ and a constant learning rate of $1\mathrm{e}{-4}$. For the Cal.\ Housing dataset we use a batch size of 256 and for the Year dataset a batch size of 1024. We repeat training for five different random initilisations (seeds).

\paragraph{Heteroskedastic Regression}
Like in the QM9 experiments, we use the same settings as for the homoskedastic model but use the loss functions and/or training modifications from the baseline methods, and tune $\beta$ for $\beta$-NLL using the same grid as in the QM9 experiments. 

\paragraph{Post-Hoc Calibration}
After fitting the homoskedastic models and the heteroskedastic models, we rescale their variance functions with a constant scale factor $\gamma >0$ using the same approach and settings as in the QM9 experiments.

\paragraph{Post-Hoc Linear Variance Estimators}
The post-hoc linear variance estimators in \cref{eq:linear_variance_prediction} are fitted using the AdamW optimizer for 200 epochs on the hold-out data using the same batch size, learning rate, and weight decay as the end-to-end trained models.

\paragraph{Post-Hoc Ensemble of Linear Variance Estimators}
Like in the QM9 experiments, we do not do any additional training or tuning for the post-hoc variance ensemble as we simply combine the variance estimators fitted using the above approach in a Gaussian mixture model using \cref{eq:variance_ensemble}.

\paragraph{Latent Representations}
For the FT-Transformer architecture, we extract four different latent representations for the different linear variance estimators. We extract the representation of the `[cls]` (prediction) token after each of the three transformer blocks, resulting in $z^{\{0,...,2\}}$, and the last-layer representation $z^3$.

\FloatBarrier
\section{Addtional Results \& Analysis}
\FloatBarrier

\FloatBarrier
\subsection{QM9}
\begin{figure}[H]
    \centering
    \begin{subfigure}[b]{0.45\textwidth}
        \centering
        \includegraphics[width=\textwidth]{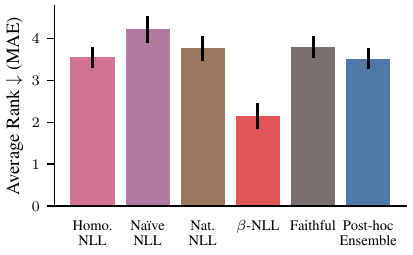}
        \caption{Test MAE Average Rankings.}
    \end{subfigure}
    \hfill 
    \begin{subfigure}[b]{0.45\textwidth}
        \centering
        \includegraphics[width=\textwidth]{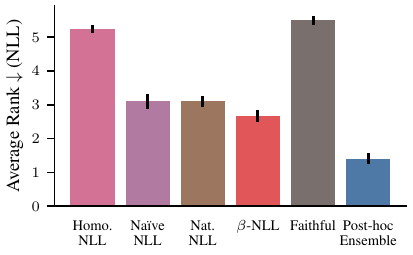}
        \caption{Test NLL Average Rankings.}
    \end{subfigure}

    \vspace{1cm} 

    \begin{subfigure}[b]{0.45\textwidth}
        \centering
        \includegraphics[width=\textwidth]{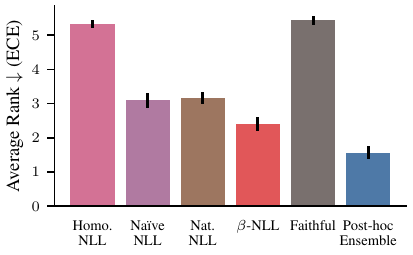}
        \caption{Test ECE Average Rankings.}
    \end{subfigure}
    \hfill
    \begin{subfigure}[b]{0.45\textwidth}
        \centering
        \includegraphics[width=\textwidth]{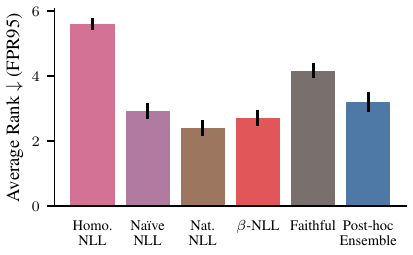}
        \caption{FPR95 (OOD detection) Average Rankings.}
    \end{subfigure}

     \vspace{1cm} 

     \begin{subfigure}[b]{0.45\textwidth}
        \centering
        \includegraphics[width=\textwidth]{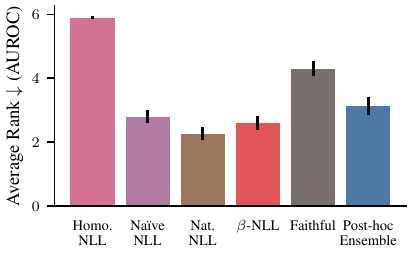}
        \caption{AUROC (OOD detection) Average Rankings.}
    \end{subfigure}
    \caption{Rankings of methods on QM9. Rankings are computed for every combination of target and seed and averaged. Error bars indicate the standard error of the mean.}
    \label{fig:app_rankings_methods}
\end{figure}

\FloatBarrier
\label{app:additional_results_qm9}
In this section, we provide additional experimental results and analysis for the methods on QM9. In ~\cref{fig:app_rankings_methods} we provide average rankings of the methods across all targets and seeds for QM9 - lower ranking is better. Here we again see that all methods are comparable in terms of MAE, except for $\beta$-NLL which achieves the best MAE on average, which either can explained by additional regularisation of the mean function, or simply due to the cross validation procedure for the $\beta$ value, which introduces a slight bias in favor of the $\beta$-NLL method. In terms of NLL and ECE on the test data, our post-hoc method achieves the best ranking, whilst several methods are comparable in terms of FPR95. We provide full tables of the OOD metrics for all methods in Table~\ref{tab:qm9_results_ood}.

\begin{figure}[H]
    \centering
    \includegraphics[width=\textwidth]{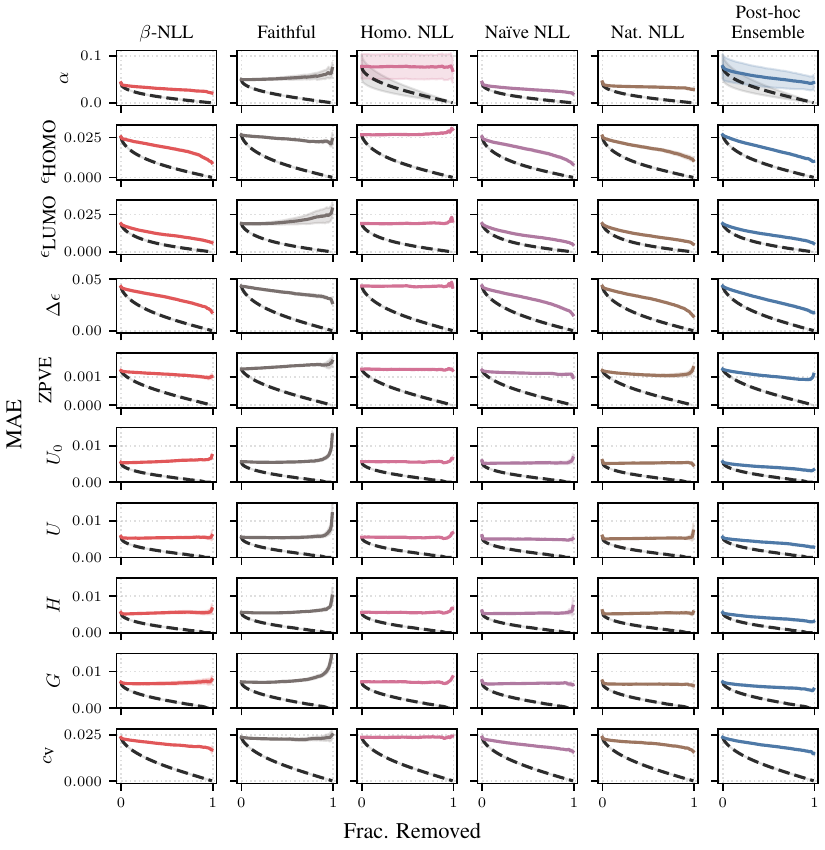}
    \caption{Calibration plots for each method for each target with average and one standard error of the mean. Plots are generated using the test set by ordering all predictions by variances and removing most uncertain points one-by-one (left side of plot with all datapoints, right side of plot with single most certain datapoint). The dotted lines are oracle lines for each model - these are created by ordering the predictions by their true residual and performing a similar procedure to obtain the full line. These indicate the calibration plot that would be obtained if the true residual was predicted. A negatively sloped line indicates correctly ranked variances on the tests set, whilst a flat line indicates a constant learnt variance prediction.}
    \label{fig:app_calibration_plots_methods}
\end{figure}

In ~\cref{fig:app_calibration_plots_methods} we provide calibration plots for each of the methods across the 10 different QM9 targets. We find that most of the methods provide well-ranked uncertainties on most of the targets - as we look at subsets of data that the model are increasingly certain about, the MAE drops as expected. The only models where this is not the case, is the homoskedastic model, which is due to the constant predicted uncertainty and the faithful model which has higher MAEs for more certain subsets. We hypothesise that this may due be to the representation issues occurring in these models, when the variance head is not allowed to affect the representation learning. Finally, we note that our post-hoc ensemble is the only method that achieves non-constant (and properly calibrated) uncertainties on the $U_o$, $U$, $H$ and $G$ targets.

\begin{figure}[H]
    \centering
    \includegraphics[width=\textwidth]{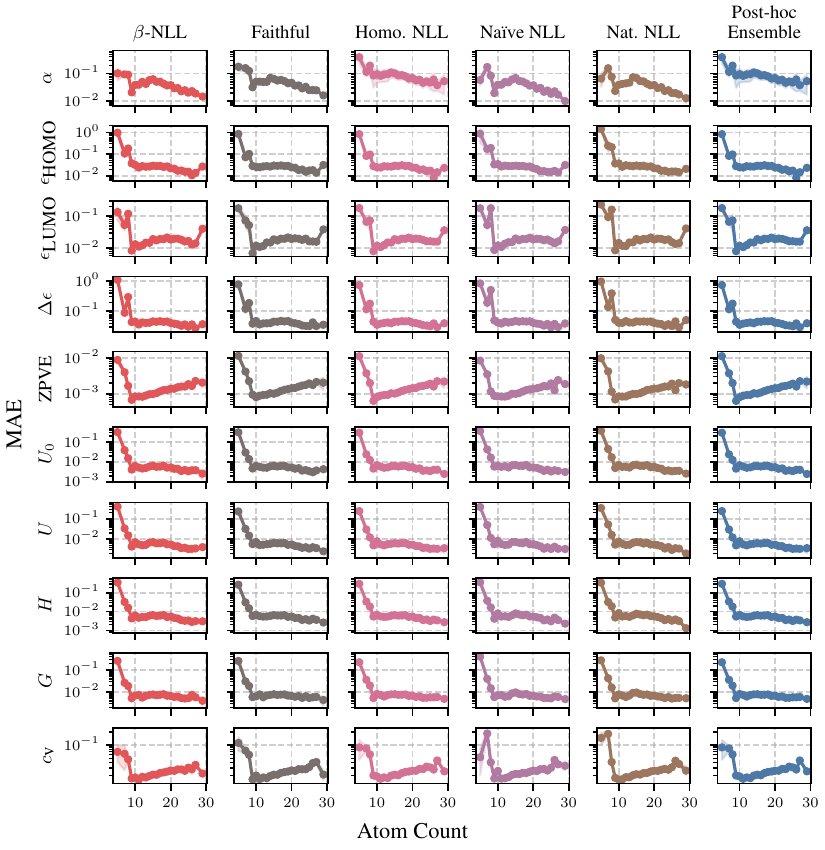}
    \caption{MAE as function of atom count of system for all methods on 10 targets from QM9. Means are presented with standard errors over three seeds.}
    \label{fig:atomc_vs_MAE}
\end{figure}

\begin{figure}[H]
    \centering
    \includegraphics[width=\textwidth]{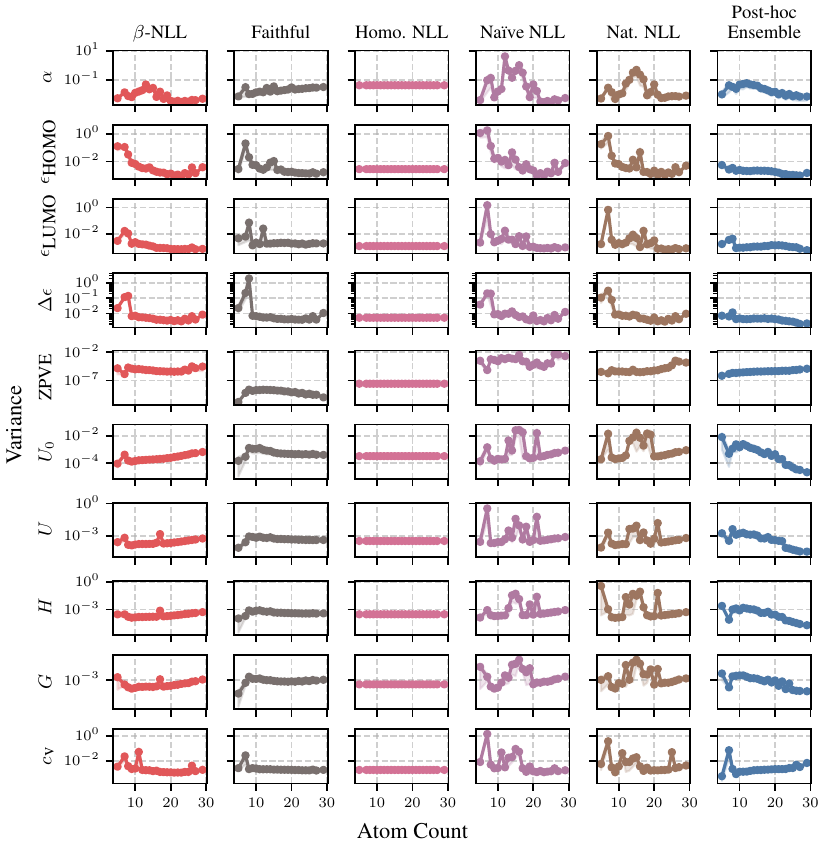}
    \caption{Variance as function of atom count of system for all methods on 10 targets from QM9. Means are presented with standard errors over three seeds.}
    \label{fig:atomc_vs_VAR}
\end{figure}

Finally, we provide an analysis of the MAE's and predictive variances on different sized atomic systems. We do this to showcase that the calibration plots are not simply an effect of the variance predictions being a sum over atom representations. We plot MAE and predicted variance as a function of atom count in Fig.~\ref{fig:atomc_vs_MAE} and Fig.~\ref{fig:atomc_vs_VAR}. Here we confirm that MAE and predicted variance are not simply monotonically increasing functions of atom count in the systems, which could give rise to the previously discussed calibration plots, but rather that the uncertainty estimates are heteroskedastic even when controlling for atomic system size.

\begin{figure}[h]
    \centering
    \includegraphics[width=0.5\textwidth]{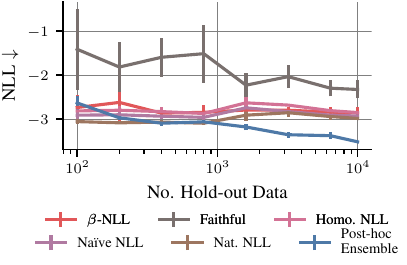}
    \caption{Effect of varying hold-out dataset size. Results are test NLL averaged over three seeds on the QM9 target $U_0$ with error bars indicating one standard error of the mean.}
    \label{fig:learning_curve_target_7}
\end{figure}

\begin{figure}[h]
    \centering
    \includegraphics[width=0.5\textwidth]{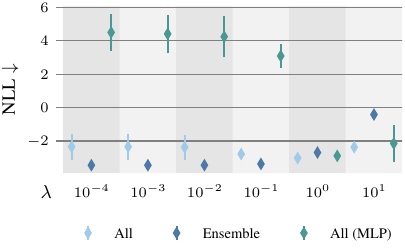}
    \caption{Ablation of variance head on the QM9 target $U$. Error bars indicate one standard error of the mean computed over seeds.}
    \label{fig:variance_head_ablation_target_8}
\end{figure}

\label{sec:ood_experiments}

\begin{table}[H]
    \caption{Out-of-distribution (OOD) results for the models trained on QM9. In-distribution (ID) data is the test split of QM9 and the OOD data is a subset of the Alchemy dataset with the same number of datapoints as the ID data. The methods' variance functions are used to discriminate between ID and OOD data and the metrics reported are the false positive rate at a 95\% true positive rate (FPR95) and the area under the receiver operating characteristic curve (AUROC). Results are averages over three seeds with best results highlighted in bold font. The standard error of the mean is reported in parenthesis in the unit of the last significant digit.}
    \centering
    \resizebox{\textwidth}{!}{%
        \begin{tabular}{lllllllllllll}
\toprule
 & \multicolumn{6}{c}{$\textrm{FPR95}\downarrow$} & \multicolumn{6}{c}{$\textrm{AUROC}\uparrow$} \\
\cmidrule(lr){2-7} \cmidrule(lr){8-13}
 & \makecell{$\textrm{Homo.}$\\$\textrm{NLL}$} & \makecell{$\textrm{Naïve}$\\$\textrm{NLL}$} & \makecell{$\textrm{Nat.}$\\$\textrm{NLL}$} & $\beta\textrm{-NLL}$ & $\textrm{Faithful}$ & \makecell{$\textrm{Post hoc}$\\$\textrm{(ours)}$} & \makecell{$\textrm{Homo.}$\\$\textrm{NLL}$} & \makecell{$\textrm{Naïve}$\\$\textrm{NLL}$} & \makecell{$\textrm{Nat.}$\\$\textrm{NLL}$} & $\beta\textrm{-NLL}$ & $\textrm{Faithful}$ & \makecell{$\textrm{Post hoc}$\\$\textrm{(ours)}$} \\
Target &  &  &  &  &  &  &  &  &  &  &  &  \\
\midrule
$\alpha$ & 0.950 {\scriptsize(0)} & 0.541 {\scriptsize(14)} & 0.682 {\scriptsize(97)} & \textbf{0.498 {\scriptsize(24)}} & 0.624 {\scriptsize(55)} & 0.696 {\scriptsize(38)} & 0.500 {\scriptsize(0)} & 0.835 {\scriptsize(2)} & 0.848 {\scriptsize(31)} & \textbf{0.853 {\scriptsize(7)}} & 0.821 {\scriptsize(33)} & 0.763 {\scriptsize(17)} \\
$\epsilon_{\textrm{HOMO}}$ & 0.950 {\scriptsize(0)} & 0.882 {\scriptsize(72)} & 0.868 {\scriptsize(94)} & 0.865 {\scriptsize(130)} & 0.992 {\scriptsize(6)} & \textbf{0.573 {\scriptsize(63)}} & 0.500 {\scriptsize(0)} & 0.680 {\scriptsize(61)} & 0.686 {\scriptsize(68)} & 0.708 {\scriptsize(101)} & 0.399 {\scriptsize(136)} & \textbf{0.867 {\scriptsize(22)}} \\
$\epsilon_{\textrm{LUMO}}$ & 0.950 {\scriptsize(0)} & 0.780 {\scriptsize(14)} & 0.741 {\scriptsize(4)} & 0.749 {\scriptsize(16)} & \textbf{0.676 {\scriptsize(44)}} & 0.756 {\scriptsize(16)} & 0.500 {\scriptsize(0)} & 0.720 {\scriptsize(3)} & 0.740 {\scriptsize(6)} & 0.741 {\scriptsize(11)} & \textbf{0.752 {\scriptsize(22)}} & 0.725 {\scriptsize(22)} \\
$\Delta \epsilon$ & 0.950 {\scriptsize(0)} & 0.938 {\scriptsize(34)} & 0.830 {\scriptsize(74)} & 0.834 {\scriptsize(129)} & 0.944 {\scriptsize(25)} & \textbf{0.476 {\scriptsize(85)}} & 0.500 {\scriptsize(0)} & 0.698 {\scriptsize(82)} & 0.763 {\scriptsize(62)} & 0.679 {\scriptsize(137)} & 0.654 {\scriptsize(65)} & \textbf{0.893 {\scriptsize(27)}} \\
$\textrm{ZPVE}$ & 0.950 {\scriptsize(0)} & 0.768 {\scriptsize(34)} & 0.716 {\scriptsize(75)} & 0.678 {\scriptsize(23)} & 0.875 {\scriptsize(40)} & \textbf{0.666 {\scriptsize(32)}} & 0.500 {\scriptsize(0)} & 0.760 {\scriptsize(17)} & 0.792 {\scriptsize(18)} & 0.809 {\scriptsize(16)} & 0.556 {\scriptsize(31)} & \textbf{0.812 {\scriptsize(13)}} \\
$U_0$ & 0.950 {\scriptsize(0)} & \textbf{0.384 {\scriptsize(24)}} & 0.388 {\scriptsize(34)} & 0.454 {\scriptsize(29)} & 0.818 {\scriptsize(28)} & 0.827 {\scriptsize(36)} & 0.500 {\scriptsize(0)} & \textbf{0.914 {\scriptsize(5)}} & 0.911 {\scriptsize(9)} & 0.895 {\scriptsize(11)} & 0.649 {\scriptsize(15)} & 0.709 {\scriptsize(32)} \\
$U$ & 0.950 {\scriptsize(0)} & 0.365 {\scriptsize(15)} & 0.360 {\scriptsize(10)} & \textbf{0.358 {\scriptsize(13)}} & 0.784 {\scriptsize(16)} & 0.814 {\scriptsize(16)} & 0.500 {\scriptsize(0)} & 0.919 {\scriptsize(5)} & \textbf{0.922 {\scriptsize(2)}} & 0.922 {\scriptsize(4)} & 0.701 {\scriptsize(21)} & 0.706 {\scriptsize(10)} \\
$H$ & 0.950 {\scriptsize(0)} & \textbf{0.353 {\scriptsize(17)}} & 0.384 {\scriptsize(37)} & 0.403 {\scriptsize(42)} & 0.784 {\scriptsize(80)} & 0.805 {\scriptsize(21)} & 0.500 {\scriptsize(0)} & \textbf{0.925 {\scriptsize(4)}} & 0.915 {\scriptsize(8)} & 0.907 {\scriptsize(11)} & 0.702 {\scriptsize(62)} & 0.723 {\scriptsize(17)} \\
$G$ & 0.950 {\scriptsize(0)} & 0.572 {\scriptsize(23)} & \textbf{0.555 {\scriptsize(13)}} & 0.600 {\scriptsize(36)} & 0.784 {\scriptsize(32)} & 0.814 {\scriptsize(16)} & 0.500 {\scriptsize(0)} & 0.821 {\scriptsize(10)} & \textbf{0.824 {\scriptsize(9)}} & 0.819 {\scriptsize(17)} & 0.681 {\scriptsize(41)} & 0.706 {\scriptsize(7)} \\
$c_{\textrm{v}}$ & 0.950 {\scriptsize(0)} & 0.483 {\scriptsize(144)} & \textbf{0.127 {\scriptsize(40)}} & 0.616 {\scriptsize(72)} & 0.744 {\scriptsize(90)} & 0.231 {\scriptsize(25)} & 0.500 {\scriptsize(0)} & 0.895 {\scriptsize(44)} & \textbf{0.973 {\scriptsize(6)}} & 0.860 {\scriptsize(24)} & 0.796 {\scriptsize(40)} & 0.958 {\scriptsize(6)} \\
\bottomrule
\end{tabular}

    }
    \label{tab:qm9_results_ood}
\end{table}

\begin{table}[H]
    \centering
    \caption{Comparison of using a linear layer on the concatenation of all representations (All), the ensemble of individual linear predictors, and an 1-layer MLP on the concatenation of all representations. $\lambda$ is the weight decay factor. Results are for the QM9 targets $U_0$ and $U$ using the PaiNN model.}
    \begin{tabular}{llllllll}
\toprule
& & \multicolumn{6}{c}{$\textrm{NLL}\downarrow$} \\
\cmidrule(lr){3-8}
 & $\lambda$ & $10^{-4}$ & $10^{-3}$ & $10^{-2}$ & $10^{-1}$ & $1$ & $10^{1}$ \\
Target & Feature &  &  &  &  &  &  \\
\midrule
\multirow[t]{3}{*}{$U_0$} & $\textrm{All}$ & \textbf{-3.25 {\scriptsize(8)}} & -3.25 {\scriptsize(9)} & -3.22 {\scriptsize(9)} & -3.11 {\scriptsize(11)} & -2.75 {\scriptsize(43)} & -2.51 {\scriptsize(32)} \\
 & $\textrm{Ensemble}$ & \textbf{-3.47 {\scriptsize(2)}} & -3.47 {\scriptsize(2)} & -3.46 {\scriptsize(2)} & -3.39 {\scriptsize(2)} & -2.73 {\scriptsize(21)} & -0.424 {\scriptsize(156)} \\
 & $\textrm{All (MLP)}$ & 6.32 {\scriptsize(59)} & 6.30 {\scriptsize(49)} & 5.64 {\scriptsize(45)} & 3.97 {\scriptsize(4)} & \textbf{-1.72 {\scriptsize(166)}} & -1.64 {\scriptsize(167)} \\
\cline{1-8}
\multirow[t]{3}{*}{$U$} & $\textrm{All}$ & -2.34 {\scriptsize(78)} & -2.34 {\scriptsize(78)} & -2.37 {\scriptsize(74)} & -2.78 {\scriptsize(28)} & \textbf{-3.02 {\scriptsize(14)}} & -2.37 {\scriptsize(27)} \\
 & $\textrm{Ensemble}$ & \textbf{-3.45 {\scriptsize(2)}} & -3.45 {\scriptsize(2)} & -3.45 {\scriptsize(1)} & -3.37 {\scriptsize(1)} & -2.69 {\scriptsize(21)} & -0.418 {\scriptsize(160)} \\
 & $\textrm{All (MLP)}$ & 4.49 {\scriptsize(110)} & 4.40 {\scriptsize(115)} & 4.24 {\scriptsize(122)} & 3.08 {\scriptsize(71)} & \textbf{-2.89 {\scriptsize(3)}} & -2.15 {\scriptsize(113)} \\
\bottomrule
\end{tabular}

\end{table}

\begin{table}[H]
    \centering
    \caption{Results of using random latent representations instead of learnt latent representations for the QM9 target $U_0$. Random representations are generated by passing the input through a randomly initialized neural network (PaiNN), extracting the representations (either the initial embeddings $z^{\mathrm{emb}}$ or the representations after the first message-passing block $z^{0}$), and computing random fourier features from these with an RBF-kernel. $\sigma$ is the length-scale of the RBF-kernel.}
    \label{tab:rff_features}
    \begin{tabular}{lllllll}
\toprule
& \multicolumn{6}{c}{$\textrm{NLL}\downarrow$} \\
\cmidrule(lr){2-7}
$\sigma$ & $10^{-2}$ & $10^{-1}$ & $1$ & $10^{1}$ & $10^{2}$ & No RFF \\
\midrule
$z^{\mathrm{emb}}$ & -2.86 {\scriptsize(5)} & -2.86 {\scriptsize(5)} & \textbf{-2.87 {\scriptsize(5)}} & -2.87 {\scriptsize(6)} & -2.82 {\scriptsize(3)} & --- \\
$z^0$ & -2.78 {\scriptsize(2)} & -2.82 {\scriptsize(3)} & \textbf{-3.02 {\scriptsize(11)}} & -2.87 {\scriptsize(7)} & -2.83 {\scriptsize(3)} & -2.95 {\scriptsize(6)} \\
$z^1$ & -2.81 {\scriptsize(2)} & -2.81 {\scriptsize(2)} & \textbf{-3.03 {\scriptsize(10)}} & -2.88 {\scriptsize(7)} & -2.83 {\scriptsize(3)} & -2.98 {\scriptsize(7)} \\
$z^2$ & -2.79 {\scriptsize(2)} & -2.77 {\scriptsize(2)} & \textbf{-3.12 {\scriptsize(1)}} & -2.88 {\scriptsize(7)} & -2.83 {\scriptsize(3)} & -2.97 {\scriptsize(8)} \\
$z^3$ & -2.76 {\scriptsize(2)} & -2.75 {\scriptsize(3)} & -2.91 {\scriptsize(6)} & -2.90 {\scriptsize(2)} & -2.78 {\scriptsize(2)} & \textbf{-2.95 {\scriptsize(7)}} \\
$z^4$ & -2.77 {\scriptsize(1)} & -2.85 {\scriptsize(4)} & -2.88 {\scriptsize(7)} & -2.88 {\scriptsize(3)} & -2.77 {\scriptsize(2)} & \textbf{-2.96 {\scriptsize(4)}} \\
$\textrm{All}$ & -2.85 {\scriptsize(5)} & -2.86 {\scriptsize(2)} & \textbf{-3.07 {\scriptsize(14)}} & -2.98 {\scriptsize(5)} & -2.90 {\scriptsize(4)} & -2.87 {\scriptsize(17)} \\
\bottomrule
\end{tabular}

\end{table}

\newpage

\FloatBarrier
\subsection{Choice of Latent Representation on QM9}
\begin{figure}[H]
    \centering
    \begin{subfigure}[b]{0.45\textwidth}
        \centering
        \includegraphics[width=\textwidth]{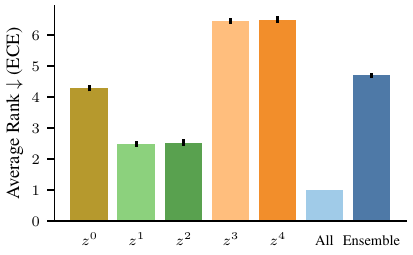}
        \caption{Test ECE Rankings for different representation usage.}
    \end{subfigure}
    \hfill 
    \begin{subfigure}[b]{0.45\textwidth}
        \centering
        \includegraphics[width=\textwidth]{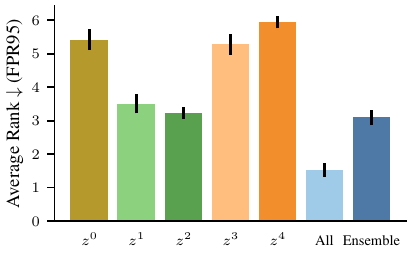}
        \caption{FPR95 (OOD detection) Rankings for different representation usage.}
    \end{subfigure}

    \vspace{1cm} 

    \begin{subfigure}[b]{0.45\textwidth}
        \centering
        \includegraphics[width=\textwidth]{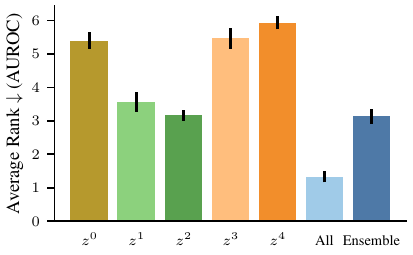}
        \caption{AUROC (OOD detection) Rankings for different representation usage.}
    \end{subfigure}
    \caption{Additional rankings when using different latent representations or combinations thereof on QM9. Rankings are computed for every combination of target and seed, and then averaged with standard error of the mean.}
    \label{app:rankings_latent_qm9}
\end{figure}

\begin{figure}[H]
    \centering
    \includegraphics[width=\columnwidth]{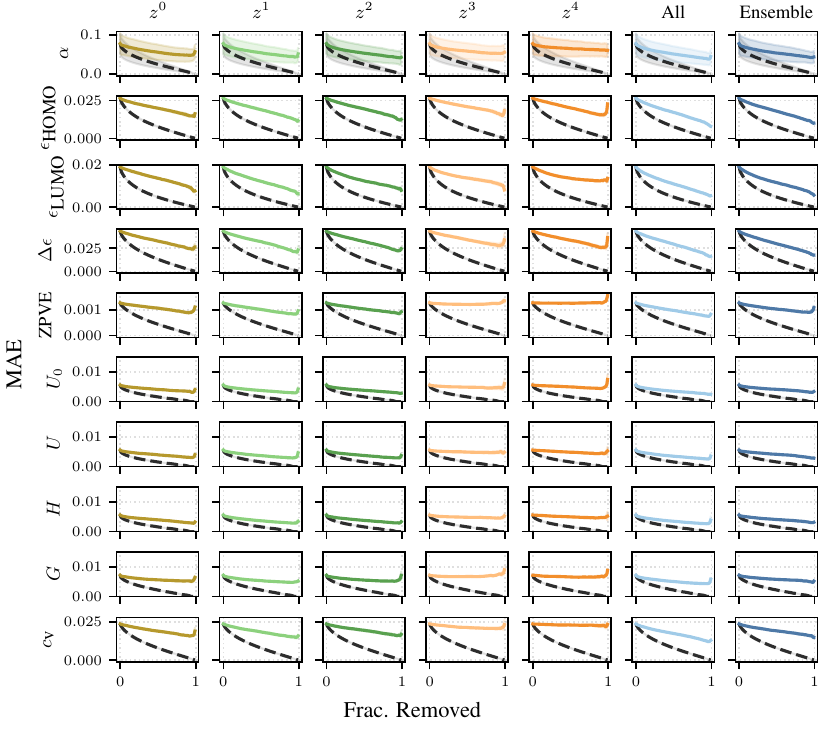}
    \caption{Calibration plots for each latent representation or combination thereof for each of the 10 targets on QM9. Each curve mean is computed over three seeds with one standard error of the mean shown. Plots are generated using the test set by ordering all predictions by variances and removing most uncertain points one-by-one (left side of plot with all datapoints, right side of plot with single most certain datapoint). The dotted lines are oracle lines for each model - these are created by ordering the predictions by their true residual and performing a similar procedure to obtain the full line. These indicate the calibration plot that would be obtained if the true residual was predicted.}
    \label{fig:calib_plot_features_qm9}
\end{figure}

\FloatBarrier
\label{app:representation_choice_appendix}
In this section of the appendix we provide additional results and analysis related to the sensitivity and ablation studies we perform to investigate the efficacy of our method and the effect of using different latent representations for the variance predictors. In \cref{app:rankings_latent_qm9} we showcase rankings of the ECE on test set, and FPR95 and AUROC on the OOD set. We additionally present the raw metrics for each target in \cref{tab:linear_var_head_representation_test,tab:linear_var_head_representation_ood}. For all metrics, we note that last layers $z^3$ and $z^4$ generally perform the worst with $z^0$ being third worst indiciating that some level of feature learning must occur to properly predict variances. In terms of NLL, we note that the ensemble is the best, but for all other metrics we see that the "All" combination performs the best - we discuss this in \cref{sec:discussion}. Finally, we also note that some seeds have very standard errors for some specific targets indicating a strong degree of overfitting; however, the ensembling procedure appears to mitigate this problem. In Fig.~\ref{fig:calib_plot_features_qm9} we showcase calibration plots for each of the versions of our method for each target. Here we again note that all versions except for those solely relying on $z^3$ or $z^4$ have heuristically good calibration.

\begin{figure}[H]
    \centering
    \begin{subfigure}[b]{0.45\textwidth}
        \centering
        \includegraphics[width=\textwidth]{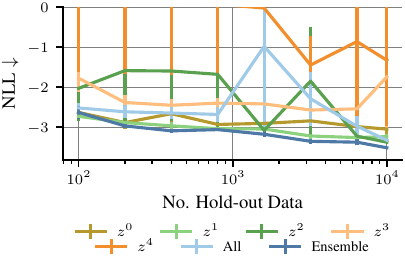}
        \caption{QM9 $U_0$ target}
    \end{subfigure}
    \hfill 
    \begin{subfigure}[b]{0.45\textwidth}
        \centering
        \includegraphics[width=\textwidth]{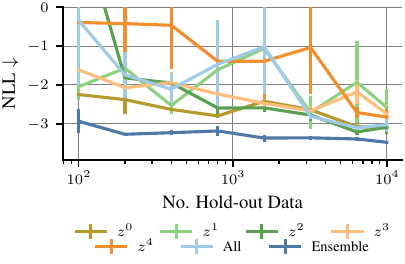}
        \caption{QM9 $U$ target}
    \end{subfigure}
    \caption{Effect of varying hold-out dataset size. Means are computed as averages across 3 repetitions and reported with one standard error of the mean.}
    \label{fig:learning_curve_appendix}
\end{figure}

\begin{figure}[H]
    \centering
    \includegraphics[width=.5\columnwidth]{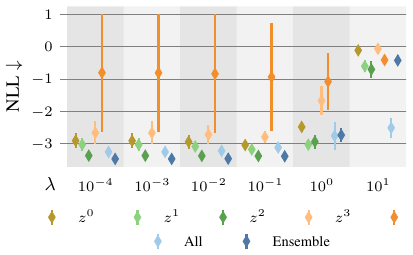}
    \caption{The effect of varying the weight decay parameter $\lambda$ when fitting the linear uncertainty estimator using different individual representations $z^{l}$, all representations, or an ensemble of the individual estimators. Results are test NLL averages over three seeds on the QM9 $U_0$ target and are plotted with one standard error.}
    \label{fig:weight_decay_qm9_target_7}
\end{figure}

We then present additional results related to and sensitivity experiments on our method. Firstly, we present an additional weight decay experiment in \cref{fig:weight_decay_qm9_target_7} where we again note that only $z^3$ and $z^4$ are very sensitive to the choice of weight decay, whilst the ensemble again is the most robust. In \cref{fig:learning_curve_appendix} we show the sensitivity of each variation of our method to number of hold-out datapoints used to fit the variance heads. Here it is clear that many of the versions are highly sensitive to the size of the hold-out dataset, except for the ensemble which consistently performs extremely well which also was noticed when comparing to the other baselines.

\begin{table}[H]
    \caption{Result of varying the latent representation $z^l$ used for variance prediction in Eq.~\eqref{eq:linear_variance_prediction}. The metrics shown are the test negative log-likelihood (NLL) and expected calibration error (ECE). Values are computed as means across three seeds with one standard error reported in parenthesis in the unit of the last significant digit. Best mean values are bolded.}
    \centering
    \resizebox{\textwidth}{!}{%
        \begin{tabular}{lllllllllllllll}
\toprule
 & \multicolumn{7}{c}{$\textrm{NLL}\downarrow$} & \multicolumn{7}{c}{$\textrm{ECE}\downarrow$} \\
\cmidrule(lr){2-8} \cmidrule(lr){9-15}
 & $z^0$ & $z^1$ & $z^2$ & $z^3$ & $z^4$ & $\textrm{All}$ & $\textrm{Ensemble}$ & $z^0$ & $z^1$ & $z^2$ & $z^3$ & $z^4$ & $\textrm{All}$ & $\textrm{Ensemble}$ \\
Target &  &  &  &  &  &  &  &  &  &  &  &  &  &  \\
\midrule
$\alpha$ & -0.542 {\scriptsize(234)} & 1.42 {\scriptsize(224)} & 0.088 {\scriptsize(1172)} & -0.775 {\scriptsize(206)} & -0.715 {\scriptsize(216)} & -0.573 {\scriptsize(82)} & \textbf{-1.04 {\scriptsize(30)}} & 0.038 {\scriptsize(9)} & 0.026 {\scriptsize(6)} & 0.022 {\scriptsize(6)} & 0.042 {\scriptsize(11)} & 0.068 {\scriptsize(18)} & \textbf{0.012 {\scriptsize(2)}} & 0.039 {\scriptsize(10)} \\
$\epsilon_{\textrm{HOMO}}$ & -1.83 {\scriptsize(2)} & -1.86 {\scriptsize(1)} & -1.78 {\scriptsize(1)} & -1.74 {\scriptsize(3)} & 1.22 {\scriptsize(199)} & -1.94 {\scriptsize(1)} & \textbf{-1.96 {\scriptsize(0)}} & 0.039 {\scriptsize(0)} & 0.033 {\scriptsize(1)} & 0.032 {\scriptsize(1)} & 0.045 {\scriptsize(0)} & 0.043 {\scriptsize(1)} & \textbf{0.016 {\scriptsize(1)}} & 0.038 {\scriptsize(1)} \\
$\epsilon_{\textrm{LUMO}}$ & -1.97 {\scriptsize(8)} & -2.14 {\scriptsize(1)} & -2.13 {\scriptsize(1)} & -2.11 {\scriptsize(2)} & -2.07 {\scriptsize(2)} & -2.24 {\scriptsize(2)} & \textbf{-2.30 {\scriptsize(1)}} & 0.054 {\scriptsize(1)} & 0.047 {\scriptsize(0)} & 0.048 {\scriptsize(1)} & 0.063 {\scriptsize(2)} & 0.061 {\scriptsize(0)} & \textbf{0.024 {\scriptsize(0)}} & 0.054 {\scriptsize(1)} \\
$\Delta \epsilon$ & -1.35 {\scriptsize(3)} & -1.37 {\scriptsize(3)} & -1.36 {\scriptsize(2)} & -1.35 {\scriptsize(3)} & 38.2 {\scriptsize(394)} & -1.44 {\scriptsize(1)} & \textbf{-1.48 {\scriptsize(0)}} & 0.033 {\scriptsize(2)} & 0.029 {\scriptsize(1)} & 0.029 {\scriptsize(1)} & 0.036 {\scriptsize(2)} & 0.035 {\scriptsize(2)} & \textbf{0.014 {\scriptsize(1)}} & 0.032 {\scriptsize(1)} \\
$\textrm{ZPVE}$ & -4.94 {\scriptsize(1)} & -4.96 {\scriptsize(1)} & -4.94 {\scriptsize(1)} & -4.83 {\scriptsize(2)} & -4.79 {\scriptsize(1)} & \textbf{-4.98 {\scriptsize(1)}} & -4.97 {\scriptsize(1)} & 0.029 {\scriptsize(1)} & 0.023 {\scriptsize(1)} & 0.025 {\scriptsize(1)} & 0.049 {\scriptsize(3)} & 0.049 {\scriptsize(2)} & \textbf{0.013 {\scriptsize(1)}} & 0.035 {\scriptsize(1)} \\
$U_0$ & -2.94 {\scriptsize(22)} & -3.08 {\scriptsize(15)} & -3.37 {\scriptsize(2)} & -2.72 {\scriptsize(30)} & -0.838 {\scriptsize(1836)} & -3.22 {\scriptsize(9)} & \textbf{-3.46 {\scriptsize(2)}} & 0.056 {\scriptsize(6)} & 0.033 {\scriptsize(1)} & 0.032 {\scriptsize(3)} & 0.111 {\scriptsize(17)} & 0.121 {\scriptsize(9)} & \textbf{0.019 {\scriptsize(2)}} & 0.069 {\scriptsize(3)} \\
$U$ & -3.06 {\scriptsize(8)} & -2.00 {\scriptsize(107)} & -3.17 {\scriptsize(11)} & -2.79 {\scriptsize(8)} & -2.67 {\scriptsize(13)} & -2.37 {\scriptsize(74)} & \textbf{-3.45 {\scriptsize(1)}} & 0.054 {\scriptsize(1)} & 0.037 {\scriptsize(1)} & 0.034 {\scriptsize(1)} & 0.104 {\scriptsize(13)} & 0.123 {\scriptsize(8)} & \textbf{0.020 {\scriptsize(1)}} & 0.070 {\scriptsize(4)} \\
$H$ & -3.12 {\scriptsize(4)} & -3.14 {\scriptsize(17)} & -3.15 {\scriptsize(16)} & -2.88 {\scriptsize(2)} & -1.86 {\scriptsize(67)} & -1.20 {\scriptsize(216)} & \textbf{-3.46 {\scriptsize(2)}} & 0.050 {\scriptsize(1)} & 0.035 {\scriptsize(2)} & 0.035 {\scriptsize(2)} & 0.106 {\scriptsize(16)} & 0.113 {\scriptsize(15)} & \textbf{0.022 {\scriptsize(1)}} & 0.067 {\scriptsize(5)} \\
$G$ & -3.02 {\scriptsize(2)} & -3.10 {\scriptsize(2)} & -2.87 {\scriptsize(8)} & -2.65 {\scriptsize(11)} & -1.16 {\scriptsize(65)} & -3.10 {\scriptsize(5)} & \textbf{-3.16 {\scriptsize(2)}} & 0.052 {\scriptsize(3)} & 0.036 {\scriptsize(2)} & 0.037 {\scriptsize(2)} & 0.100 {\scriptsize(12)} & 0.107 {\scriptsize(10)} & \textbf{0.023 {\scriptsize(2)}} & 0.067 {\scriptsize(3)} \\
$c_{\textrm{v}}$ & -1.96 {\scriptsize(3)} & -2.01 {\scriptsize(1)} & -1.98 {\scriptsize(1)} & -1.84 {\scriptsize(1)} & -1.32 {\scriptsize(54)} & \textbf{-2.07 {\scriptsize(1)}} & -2.04 {\scriptsize(1)} & 0.032 {\scriptsize(3)} & 0.027 {\scriptsize(1)} & 0.032 {\scriptsize(5)} & 0.067 {\scriptsize(4)} & 0.065 {\scriptsize(3)} & \textbf{0.013 {\scriptsize(2)}} & 0.044 {\scriptsize(2)} \\
\bottomrule
\end{tabular}

    }
    \label{tab:linear_var_head_representation_test}
\end{table}

\begin{table}[H]
    \centering
    \caption{Effect of the representation $z^l$ used for linear variance prediction on OOD metrics. Values are computed as means across three seeds with one standard error reported in parenthesis in the unit of the last significant digit. Best mean values are bolded.}
    \resizebox{\textwidth}{!}{%
        \begin{tabular}{lllllllllllllll}
\toprule
 & \multicolumn{7}{c}{$\textrm{FPR95}\downarrow$} & \multicolumn{7}{c}{$\textrm{AUROC}\uparrow$} \\
\cmidrule(lr){2-8} \cmidrule(lr){9-15}
 & $z^0$ & $z^1$ & $z^2$ & $z^3$ & $z^4$ & $\textrm{All}$ & $\textrm{Ensemble}$ & $z^0$ & $z^1$ & $z^2$ & $z^3$ & $z^4$ & $\textrm{All}$ & $\textrm{Ensemble}$ \\
Target &  &  &  &  &  &  &  &  &  &  &  &  &  &  \\
\midrule
$\alpha$ & 0.878 {\scriptsize(38)} & 0.776 {\scriptsize(8)} & 0.702 {\scriptsize(50)} & 0.761 {\scriptsize(15)} & 0.757 {\scriptsize(30)} & \textbf{0.587 {\scriptsize(61)}} & 0.696 {\scriptsize(38)} & 0.611 {\scriptsize(70)} & 0.725 {\scriptsize(20)} & 0.779 {\scriptsize(26)} & 0.751 {\scriptsize(16)} & 0.731 {\scriptsize(11)} & \textbf{0.840 {\scriptsize(29)}} & 0.763 {\scriptsize(17)} \\
$\epsilon_{\textrm{HOMO}}$ & 0.811 {\scriptsize(22)} & 0.658 {\scriptsize(69)} & 0.642 {\scriptsize(79)} & 0.691 {\scriptsize(129)} & 0.868 {\scriptsize(58)} & \textbf{0.464 {\scriptsize(216)}} & 0.573 {\scriptsize(63)} & 0.679 {\scriptsize(34)} & 0.817 {\scriptsize(29)} & 0.815 {\scriptsize(30)} & 0.778 {\scriptsize(71)} & 0.650 {\scriptsize(132)} & \textbf{0.867 {\scriptsize(71)}} & 0.867 {\scriptsize(22)} \\
$\epsilon_{\textrm{LUMO}}$ & 0.859 {\scriptsize(5)} & 0.769 {\scriptsize(7)} & 0.755 {\scriptsize(9)} & 0.808 {\scriptsize(51)} & 0.829 {\scriptsize(21)} & \textbf{0.621 {\scriptsize(19)}} & 0.756 {\scriptsize(16)} & 0.657 {\scriptsize(12)} & 0.716 {\scriptsize(43)} & 0.725 {\scriptsize(14)} & 0.663 {\scriptsize(45)} & 0.600 {\scriptsize(27)} & \textbf{0.823 {\scriptsize(25)}} & 0.725 {\scriptsize(22)} \\
$\Delta \epsilon$ & 0.849 {\scriptsize(22)} & 0.816 {\scriptsize(30)} & 0.529 {\scriptsize(87)} & 0.586 {\scriptsize(100)} & 0.684 {\scriptsize(122)} & \textbf{0.410 {\scriptsize(75)}} & 0.476 {\scriptsize(85)} & 0.668 {\scriptsize(32)} & 0.739 {\scriptsize(30)} & 0.871 {\scriptsize(33)} & 0.843 {\scriptsize(39)} & 0.814 {\scriptsize(45)} & \textbf{0.913 {\scriptsize(20)}} & 0.893 {\scriptsize(27)} \\
$\textrm{ZPVE}$ & 0.805 {\scriptsize(19)} & 0.746 {\scriptsize(15)} & 0.718 {\scriptsize(33)} & 0.874 {\scriptsize(16)} & 0.790 {\scriptsize(41)} & 0.712 {\scriptsize(3)} & \textbf{0.666 {\scriptsize(32)}} & 0.741 {\scriptsize(21)} & 0.782 {\scriptsize(11)} & 0.783 {\scriptsize(14)} & 0.602 {\scriptsize(21)} & 0.666 {\scriptsize(38)} & \textbf{0.821 {\scriptsize(14)}} & 0.812 {\scriptsize(13)} \\
$U_0$ & 0.850 {\scriptsize(29)} & 0.704 {\scriptsize(26)} & 0.710 {\scriptsize(29)} & 0.900 {\scriptsize(33)} & 0.911 {\scriptsize(21)} & \textbf{0.620 {\scriptsize(16)}} & 0.827 {\scriptsize(36)} & 0.685 {\scriptsize(16)} & 0.780 {\scriptsize(21)} & 0.779 {\scriptsize(22)} & 0.590 {\scriptsize(65)} & 0.627 {\scriptsize(41)} & \textbf{0.845 {\scriptsize(6)}} & 0.709 {\scriptsize(32)} \\
$U$ & 0.839 {\scriptsize(29)} & 0.677 {\scriptsize(22)} & 0.747 {\scriptsize(24)} & 0.917 {\scriptsize(16)} & 0.893 {\scriptsize(16)} & \textbf{0.628 {\scriptsize(28)}} & 0.814 {\scriptsize(16)} & 0.682 {\scriptsize(19)} & 0.785 {\scriptsize(11)} & 0.765 {\scriptsize(11)} & 0.567 {\scriptsize(52)} & 0.616 {\scriptsize(28)} & \textbf{0.847 {\scriptsize(7)}} & 0.706 {\scriptsize(10)} \\
$H$ & 0.815 {\scriptsize(28)} & 0.699 {\scriptsize(16)} & 0.747 {\scriptsize(43)} & 0.877 {\scriptsize(23)} & 0.899 {\scriptsize(28)} & \textbf{0.664 {\scriptsize(7)}} & 0.805 {\scriptsize(21)} & 0.701 {\scriptsize(20)} & 0.777 {\scriptsize(9)} & 0.762 {\scriptsize(25)} & 0.627 {\scriptsize(40)} & 0.584 {\scriptsize(59)} & \textbf{0.829 {\scriptsize(2)}} & 0.723 {\scriptsize(17)} \\
$G$ & 0.737 {\scriptsize(42)} & 0.729 {\scriptsize(21)} & 0.775 {\scriptsize(19)} & 0.887 {\scriptsize(48)} & 0.908 {\scriptsize(10)} & \textbf{0.713 {\scriptsize(43)}} & 0.814 {\scriptsize(16)} & 0.742 {\scriptsize(36)} & 0.756 {\scriptsize(6)} & 0.746 {\scriptsize(6)} & 0.624 {\scriptsize(70)} & 0.585 {\scriptsize(18)} & \textbf{0.811 {\scriptsize(10)}} & 0.706 {\scriptsize(7)} \\
$c_{\textrm{v}}$ & 0.681 {\scriptsize(2)} & 0.639 {\scriptsize(3)} & 0.348 {\scriptsize(97)} & 0.610 {\scriptsize(138)} & 0.724 {\scriptsize(81)} & 0.268 {\scriptsize(119)} & \textbf{0.231 {\scriptsize(25)}} & 0.820 {\scriptsize(7)} & 0.841 {\scriptsize(9)} & 0.932 {\scriptsize(21)} & 0.851 {\scriptsize(49)} & 0.797 {\scriptsize(44)} & 0.952 {\scriptsize(20)} & \textbf{0.958 {\scriptsize(6)}} \\
\bottomrule
\end{tabular}

    }
    \label{tab:linear_var_head_representation_ood}
\end{table}

\FloatBarrier
\subsection{Last-Layer Representation Collapse \& Geometry}
\begin{figure}[h]
    \centering
    \includegraphics[width=\textwidth]{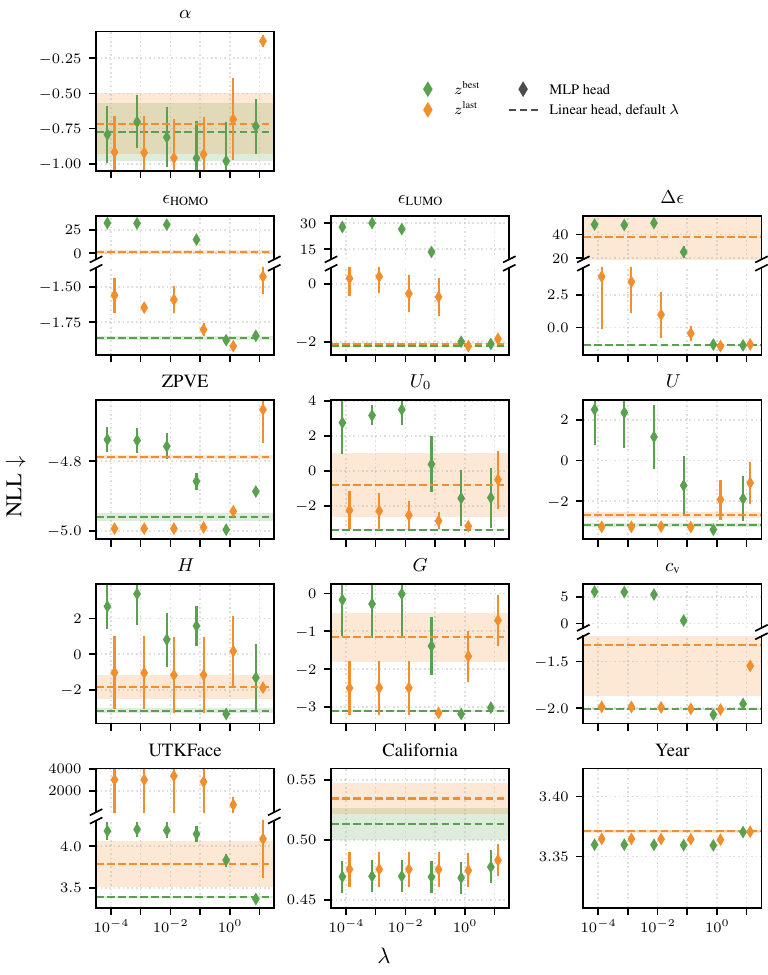}
    \caption{Test \ac{nll} of a 1-layer \ac{mlp} variance head fitted on held-out data against its weight decay $\lambda$, and of the linear head at the default $\lambda$, on the best intermediate ($z^{\mathrm{best}}$) and last-layer ($z^{\mathrm{last}}$) representation. Results are mean and one standard error over three seeds (five for \textsc{california} and \textsc{year}).}
    \label{fig:variance_head_sweep}
\end{figure}

\FloatBarrier
\label{app:last_layer}
In \cref{fig:latent_ablation}, we consistently saw that last-layer representations from mean-only networks were suboptimal for fitting linear variance predictors post hoc. 
In \cref{sec:feature_collapse}, we hypothesised that this could be either due to representation collapse or an unfavourable geometry in these representations. We now investigate this in more detail by (i) fitting \acp{mlp} to the last-layer representations, $z^{\mathrm{last}}$, and (ii) fitting \acp{mlp} to the intermediate representations that were most useful for linear variance prediction, $z^{\mathrm{best}}$. The results of this experiment are shown in \cref{fig:variance_head_sweep}. The choice of $z^{\mathrm{best}}$ is based on the test \ac{nll} of the linear predictors at their default weight decay, while the weight decay is swept for each \ac{mlp}.

In \cref{fig:variance_head_sweep}, we see that linear variance predictors fitted on $z^{\mathrm{best}}$ are preferable to those fitted on $z^{\mathrm{last}}$ on every target, as already established. However, this performance difference is recovered on most targets (fully on ten of the 13 targets and partially on $U_0$) if we instead fit an \ac{mlp} on $z^{\mathrm{last}}$ with a suitable weight decay, $\lambda$, indicating that the poor performance of $z^{\mathrm{last}}$ is often not due to representation collapse, but instead due to the variance-related information not being linearly accessible in these representations. This conclusion is further supported by the fact that \acp{mlp} fitted on $z^{\mathrm{best}}$ do not generally outperform those fitted on $z^{\mathrm{last}}$. Only on $H$ and UTKFace does an \ac{mlp} on $z^{\mathrm{last}}$ not close the gap, which is consistent with genuine representation collapse on these two targets. Interestingly, \acp{mlp} fitted on either representation do not generally outperform the linear predictors on $z^{\mathrm{best}}$ in a substantial way, indicating that all variance-related information is linearly accessible within the model, i.e., in intermediate representations.

From a practical perspective, \cref{fig:variance_head_sweep} shows that the \ac{mlp} variance predictors are highly sensitive to the weight decay factor, $\lambda$, and can fail catastrophically if $\lambda$ is not set appropriately. Furthermore, the \acp{mlp} considered here require $\sim\!10^{2}\times$ more parameters than, e.g., our \ac{ensemble} approach, which is still favourable to the single linear variance predictors.

\FloatBarrier
\subsection{OMol25}
\FloatBarrier
\label{app:additional_results_omol25}

\begin{figure}[H]
    \centering
    \includegraphics[width=\textwidth]{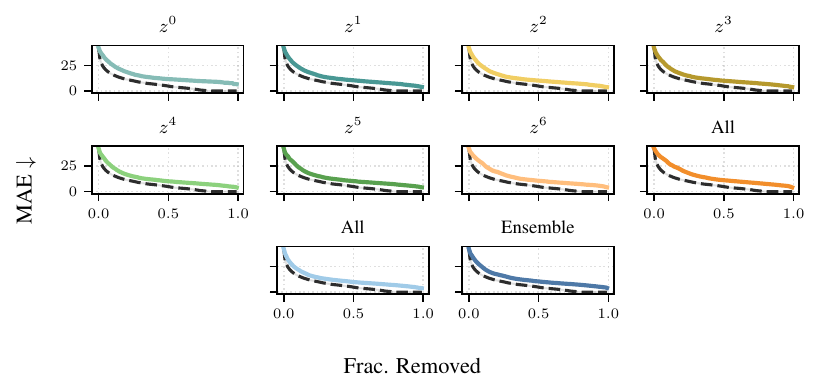}
    \caption{Calibration plots for each representation using AllScAIP on OMol25. Each curve mean is computed over three seeds with one standard error of the mean shown. Plots are generated using the test set by ordering all predictions by variances and removing most uncertain points one-by-one (left side of plot with all datapoints, right side of plot with single most certain datapoint). The dotted lines are oracle lines for each model - these are created by ordering the predictions by their true residual and performing a similar procedure to obtain the full line. These indicate the calibration plot that would be obtained if the true residual was predicted.}
    \label{fig:calib_plot_allscaip}
\end{figure}

\begin{figure}[H]
    \centering
    \includegraphics[width=.5\columnwidth]{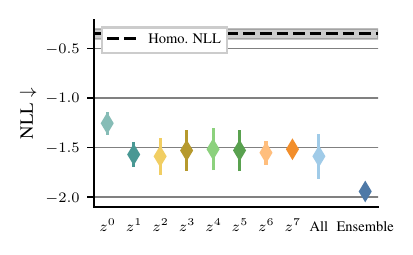}
    \caption{OMol25 test NLL when the fitting the linear variance estimator using different individual representations $z^l$, all representations, or an ensemble of the individual estimators. Results are averages over three seeds and subsets of the full data. Error bars indicate the standard error of the mean.}
    \label{fig:omol_representations_allscaip}
\end{figure}

\begin{figure}[H]
    \centering
    \includegraphics[width=\textwidth]{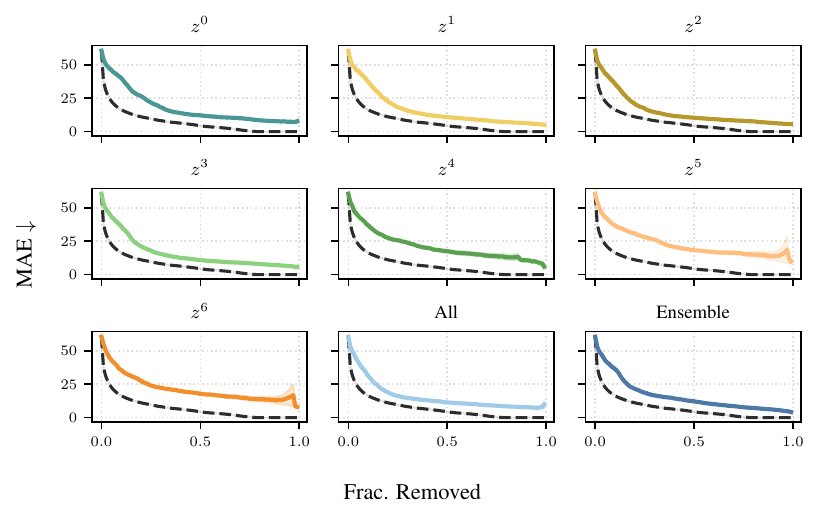}
    \caption{Calibration plots for each representation using UMA on OMol25. Each curve mean is computed over three seeds with one standard error of the mean shown. Plots are generated using the test set by ordering all predictions by variances and removing most uncertain points one-by-one (left side of plot with all datapoints, right side of plot with single most certain datapoint). The dotted lines are oracle lines for each model - these are created by ordering the predictions by their true residual and performing a similar procedure to obtain the full line. These indicate the calibration plot that would be obtained if the true residual was predicted.}
    \label{fig:calib_plot_uma}
\end{figure}

In this part of the appendix we provide additional results and plots related to the OMol25 experiments. In \cref{fig:calib_plot_uma} and \cref{fig:calib_plot_allscaip} we provide calibration plots for UMA and AllScAIP variance heads respectively. In UMA we clearly again see how later representations provide worse calibrated uncertanties than earlier layers, with "All" and the ensemble performing even better. Interestingly we find that this pattern does not repeat for AllScAIP as corroborated by the calibration plots, where "All" performs the best, although with minor differences to the other method versions. This is further corroborated by \cref{fig:omol_representations_allscaip} where we showcase the NLL when using each different latent representation or combination thereof. Here we can see that several representations provide very strong NLLs, indicating that perhaps the latent representations throughout AllScAIP are less collapsed or that directions necessary for variance predictions remain linearly separable in Transformer architectures.

\begin{table}[H]
    \centering
    \caption{Additional Baselines for the UMA and AllScAIP experiments. BLR is Bayesian linear regression on the last-layer representations and LLLA is last-layer Laplace with a diagonal covariance structure. We fit both of these using the hold-out data and optimize the prior variance and observation noise using the marginal likelihood for both.}
    \begin{tabular}{lcccc}
\toprule
 & \multicolumn{2}{c}{$\textrm{MAE}\downarrow\;(\times 10^3)$} & \multicolumn{2}{c}{$\textrm{NLL}\downarrow$} \\
\cmidrule(lr){2-3} \cmidrule(lr){4-5}
 & $\textrm{BLR}$ & $\textrm{LLLA}$ & $\textrm{BLR}$ & $\textrm{LLLA}$ \\
\midrule
$\textrm{UMA}$ & \textbf{87.6 {\scriptsize(53)}} & 61.3 {\scriptsize(9)} & 0.261 {\scriptsize(18)} & 0.265 {\scriptsize(9)} \\
$\textrm{AllScAIP}$ & 137 {\scriptsize(30)} & \textbf{42.8 {\scriptsize(1)}} & \textbf{0.138 {\scriptsize(22)}} & \textbf{-0.403 {\scriptsize(87)}} \\
\bottomrule
\end{tabular}

\end{table}

\FloatBarrier
\subsection{ResNet50 on UTKFace}
\FloatBarrier
\label{app:utkface}

\begin{table}[H]
    \centering
    \caption{Post-hoc linear variance prediction using the different latent representations $z^{0,\dots,4}$ for age estimation on the UTKFace dataset with ResNet50. Convolutional feature maps are average-pooled across the spatial  dimensions to make the dimensionality of the latent representations more manageable.}
    \resizebox{\textwidth}{!}{\begin{tabular}{lllllllllllllll}
\toprule
 & \multicolumn{7}{c}{$\textrm{NLL}\downarrow$} & \multicolumn{7}{c}{$\textrm{ECE}\downarrow$} \\
\cmidrule(lr){2-8} \cmidrule(lr){9-15}
 & $z^0$ & $z^1$ & $z^2$ & $z^3$ & $z^4$ & $\textrm{All}$ & $\textrm{Ensemble}$ & $z^0$ & $z^1$ & $z^2$ & $z^3$ & $z^4$ & $\textrm{All}$ & $\textrm{Ensemble}$ \\
Pooling &  &  &  &  &  &  &  &  &  &  &  &  &  &  \\
\midrule
\\
Spatial & 3.40 {\scriptsize(0)} & 3.39 {\scriptsize(0)} & 3.39 {\scriptsize(1)} & 3.79 {\scriptsize(28)} & 3.79 {\scriptsize(28)} & 3.48 {\scriptsize(3)} & \textbf{3.25 {\scriptsize(0)}} & 0.044 {\scriptsize(2)} & 0.042 {\scriptsize(2)} & 0.033 {\scriptsize(3)} & 0.022 {\scriptsize(3)} & 0.022 {\scriptsize(3)} & \textbf{0.017 {\scriptsize(4)}} & 0.032 {\scriptsize(2)} \\
\bottomrule
\end{tabular}
}
\end{table}

\FloatBarrier
\subsection{FT-Transformer on California Housing and Year Datasets}
\FloatBarrier
\label{app:tabular}

\begin{table}[H]
    \centering
    \caption{Post-hoc linear variance prediction using the different latent representations $z^{0,\dots,3}$ for for heteroskedastic regression on the tabular datasets California Housing and Year with the FT-Transformer model}
    \resizebox{\textwidth}{!}{\begin{tabular}{lllllllllllll}
\toprule
 & \multicolumn{6}{c}{$\textrm{NLL}\downarrow$} & \multicolumn{6}{c}{$\textrm{ECE}\downarrow$} \\
\cmidrule(lr){2-7} \cmidrule(lr){8-13}
 & $z^0$ & $z^1$ & $z^2$ & $z^3$ & $\textrm{All}$ & $\textrm{Ensemble}$ & $z^0$ & $z^1$ & $z^2$ & $z^3$ & $\textrm{All}$ & $\textrm{Ensemble}$ \\
Dataset &  &  &  &  &  &  &  &  &  &  &  &  \\
\midrule
Cal. Housing & 0.645 {\scriptsize(8)} & 0.539 {\scriptsize(15)} & 0.513 {\scriptsize(13)} & 0.534 {\scriptsize(13)} & \textbf{0.490 {\scriptsize(14)}} & 0.519 {\scriptsize(13)} & 0.062 {\scriptsize(2)} & 0.043 {\scriptsize(1)} & 0.039 {\scriptsize(1)} & 0.043 {\scriptsize(1)} & \textbf{0.034 {\scriptsize(2)}} & 0.046 {\scriptsize(1)} \\
Year & 3.47 {\scriptsize(0)} & 3.39 {\scriptsize(0)} & 3.37 {\scriptsize(0)} & 3.37 {\scriptsize(0)} & 3.36 {\scriptsize(0)} & \textbf{3.36 {\scriptsize(0)}} & 0.041 {\scriptsize(3)} & 0.037 {\scriptsize(4)} & \textbf{0.037 {\scriptsize(4)}} & 0.038 {\scriptsize(4)} & 0.037 {\scriptsize(4)} & 0.038 {\scriptsize(4)} \\
\bottomrule
\end{tabular}
}
\end{table}

\FloatBarrier
\subsection{Uncalibrated Baselines}
\FloatBarrier
\label{app:uncalibrated_baselines}
Here we provide negative log-likelihoods for the baseline models before they are recalibrated with variance rescaling (temperature scaling) on the hold-out data. We note that the QM9 and UTKFace baselines have extremely poor likelihoods before rescaling, highlighting the need to recalibrate end-to-end trained mean-variance networks. We do, however, not observe the same degree of degraded performance for the tabular baselines. We hypothesize that this is because these models are early-stopped extremely quickly (based on the performance of the mean) before they reach the interpolating stage and the variance function collapses, i.e., we are not in the benign overfitting regime. 
\begin{table}[H]
    \centering
    \caption{Negative log-likelihood (NLL) for the baseline methods on QM9 before recalibrating them on the hold-out data by fitting a constant variance scale factor.}
    \begin{tabular}{llllll}
\toprule
 &  & \multicolumn{4}{c}{$\textrm{NLL}\downarrow$} \\
\cmidrule(lr){3-6}
 &  & \makecell{$\textrm{Naïve}$\\$\textrm{NLL}$} & \makecell{$\textrm{Nat.}$\\$\textrm{NLL}$} & $\beta\textrm{-NLL}$ & $\textrm{Faithful}$ \\
 &  &  &  &  &  \\
\midrule
\multirow{10}{*}{\rotatebox{90}{QM9}} & $\alpha$ & 127 {\scriptsize(41)} & 11.2 {\scriptsize(2)} & 11.6 {\scriptsize(59)} & 47.7 {\scriptsize(94)} \\
 & $\epsilon_{\textrm{HOMO}}$ & 281 {\scriptsize(173)} & 264 {\scriptsize(100)} & 62.0 {\scriptsize(142)} & 164 {\scriptsize(72)} \\
 & $\epsilon_{\textrm{LUMO}}$ & 212 {\scriptsize(47)} & 346 {\scriptsize(55)} & 81.4 {\scriptsize(34)} & 417 {\scriptsize(88)} \\
 & $\Delta \epsilon$ & 172 {\scriptsize(47)} & 169 {\scriptsize(53)} & 66.0 {\scriptsize(232)} & 85.3 {\scriptsize(264)} \\
 & $\textrm{ZPVE}$ & -4.49 {\scriptsize(3)} & -4.56 {\scriptsize(7)} & -4.38 {\scriptsize(12)} & -4.12 {\scriptsize(4)} \\
 & $U_0$ & 28.5 {\scriptsize(35)} & 20.2 {\scriptsize(15)} & 30.1 {\scriptsize(58)} & 27.2 {\scriptsize(17)} \\
 & $U$ & 27.8 {\scriptsize(69)} & 25.4 {\scriptsize(26)} & 31.4 {\scriptsize(67)} & 24.7 {\scriptsize(50)} \\
 & $H$ & 42.9 {\scriptsize(203)} & 25.8 {\scriptsize(76)} & 28.9 {\scriptsize(15)} & 26.8 {\scriptsize(57)} \\
 & $G$ & 31.1 {\scriptsize(27)} & 27.1 {\scriptsize(72)} & 15.5 {\scriptsize(57)} & 27.4 {\scriptsize(53)} \\
 & $c_{\textrm{v}}$ & 31.5 {\scriptsize(57)} & 10.4 {\scriptsize(6)} & 18.5 {\scriptsize(43)} & 21.3 {\scriptsize(37)} \\
\midrule
\multicolumn{2}{l}{UTKFace} & 27.4 {\scriptsize(149)} & 54.9 {\scriptsize(285)} & 12.1 {\scriptsize(6)} & 104 {\scriptsize(69)} \\
\midrule
\multicolumn{2}{l}{Cal. Housing} & 0.679 {\scriptsize(75)} & 0.681 {\scriptsize(102)} & 0.449 {\scriptsize(3)} & 0.494 {\scriptsize(6)} \\
\multicolumn{2}{l}{Year} & 3.43 {\scriptsize(1)} & 3.46 {\scriptsize(3)} & 3.37 {\scriptsize(0)} & 3.38 {\scriptsize(0)} \\
\bottomrule
\end{tabular}

    \label{tab:uncalibrated_baselines_new}
\end{table}

\FloatBarrier
\subsection{Isotonic Regression}
\FloatBarrier
\label{app:isotonic}
We test using isotonic regression as a calibration method for each of the end-to-end trained mean-variance networks as an alternative to temperature scaling. One might hypothesize that a more flexible calibration method could boost the NLL of the end-to-end methods, but we generally do not find this to be the case - in fact on some datasets this actually overfits, leading to poor generalisation. We fit the isotonic regressor on the hold-out dataset using Sci-Kit Learn's \citep{pedregosa2011scikit} IsotonicRegressor class, where the input is the uncalibrated variance predictions on the hold-out points, and the target of the isotonic regressor are the true (squared) residuals on the hold-out points. At test time, predicted variances are then passed through the isotonic regressor to obtain the calibrated residual.
\begin{table}[H]
    \centering
    \caption{Recalibration of the baselines on the hold-out data using isotonic regression.}
    \begin{tabular}{llllll}
\toprule
 &  & \multicolumn{4}{c}{$\textrm{NLL}\downarrow$} \\
\cmidrule(lr){3-6}
 &  & \makecell{$\textrm{Naïve}$\\$\textrm{NLL}$} & \makecell{$\textrm{Nat.}$\\$\textrm{NLL}$} & $\beta\textrm{-NLL}$ & $\textrm{Faithful}$ \\
 &  &  &  &  &  \\
\midrule
\multirow{10}{*}{\rotatebox{90}{QM9}} & $\alpha$ & -1.37 {\scriptsize(5)} & -1.34 {\scriptsize(1)} & -1.40 {\scriptsize(1)} & 0.049 {\scriptsize(825)} \\
 & $\epsilon_{\textrm{HOMO}}$ & -1.93 {\scriptsize(3)} & -1.90 {\scriptsize(3)} & -1.91 {\scriptsize(2)} & -1.76 {\scriptsize(1)} \\
 & $\epsilon_{\textrm{LUMO}}$ & -2.23 {\scriptsize(2)} & -2.11 {\scriptsize(5)} & -2.27 {\scriptsize(2)} & -2.00 {\scriptsize(2)} \\
 & $\Delta \epsilon$ & -1.44 {\scriptsize(1)} & -1.44 {\scriptsize(0)} & -1.44 {\scriptsize(1)} & -1.33 {\scriptsize(1)} \\
 & $\textrm{ZPVE}$ & -4.96 {\scriptsize(1)} & -4.91 {\scriptsize(2)} & -4.96 {\scriptsize(1)} & -4.83 {\scriptsize(1)} \\
 & $U_0$ & -2.94 {\scriptsize(3)} & -2.97 {\scriptsize(2)} & -2.89 {\scriptsize(3)} & -2.85 {\scriptsize(2)} \\
 & $U$ & -2.88 {\scriptsize(11)} & -3.11 {\scriptsize(3)} & -2.97 {\scriptsize(7)} & -2.82 {\scriptsize(1)} \\
 & $H$ & -2.78 {\scriptsize(27)} & -3.06 {\scriptsize(9)} & -2.97 {\scriptsize(9)} & -2.85 {\scriptsize(4)} \\
 & $G$ & -2.81 {\scriptsize(4)} & -2.84 {\scriptsize(8)} & -2.88 {\scriptsize(6)} & -2.75 {\scriptsize(2)} \\
 & $c_{\textrm{v}}$ & -2.01 {\scriptsize(0)} & -1.99 {\scriptsize(1)} & -2.02 {\scriptsize(1)} & -1.92 {\scriptsize(0)} \\
\midrule
\multicolumn{2}{l}{UTKFace} & 3.86 {\scriptsize(47)} & 9.23 {\scriptsize(594)} & 6.57 {\scriptsize(196)} & 4.25 {\scriptsize(48)} \\
\midrule
\multicolumn{2}{l}{Cal. Housing} & 0.415 {\scriptsize(27)} & 0.551 {\scriptsize(166)} & 0.441 {\scriptsize(21)} & 0.505 {\scriptsize(8)} \\
\multicolumn{2}{l}{Year} & 3.37 {\scriptsize(1)} & 3.37 {\scriptsize(1)} & 3.35 {\scriptsize(0)} & 3.38 {\scriptsize(0)} \\
\bottomrule
\end{tabular}

\end{table}

\FloatBarrier
\subsection{Conformal Prediction}
\FloatBarrier
In \cref{tab:width}, we provide average marginal coverage, and average and median interval widths at different nominal levels for all methods as well as \acl{scp} (\acs{scp}) and \acl{lacp} (\acs{lacp}). \ac{scp} and \ac{lacp} are obtained by conformalising the homoskedastic baseline and the naïve Gaussian \ac{nll} baseline, respectively, with the held-out data used to recalibrate the baselines and to fit the post-hoc methods. The interval widths in \cref{tab:width} are computed relative to the interval width of \ac{scp} due to the targets being on different scales (e.g., a width of $1.5$ means $1.5 \times$ the \ac{scp} width). We see in \cref{tab:width} that the conformal methods provide the expected coverages, while most mean-variance networks fail to provide the same. We do, however, note that it is possible to conformalise all methods to provide coverage guarantees if that is of interest to a practitioner. Interestingly, we also see that our \ac{ensemble} approach overcovers at all but the $99\%$ level (due to its heavy tails), while \ac{all} provides relatively sharper intervals, but as a result undercovers at all but the $80\%$ level.

In \cref{tab:intervals}, we provide the interval score that measures the trade-off between coverage and interval width. Loosely speaking, the interval score rewards narrow intervals for correct predictions and large intervals for incorrect predictions. In \cref{tab:intervals}, we rank each method by its interval score on each target and average across all targets for each nominal level. Here, we note that \ac{all} is ranked best overall and \ac{ensemble} is ranked third (best at the $99\%$ level), indicating that our method is a strong compromise between narrow interval widths and good coverage.

\label{app:conformal}
\begin{table}[h]
    \centering
    \caption{Marginal coverage and mean \& median width at different nominal levels ($1-\alpha$). Widths are computed relative to \ac{scp} and results are averaged across all 13 targets (ten \textsc{qm}{\footnotesize 9}, \textsc{utkface}, \textsc{california}, and \textsc{year}). $^*$Mean \& median widths for the post-hoc methods at $\alpha$s that match the desired empirical coverage (not possible in practice but included for width comparability at the same coverages).}
    \label{tab:width}
    \resizebox{\textwidth}{!}{%
    \begin{tabular}{@{}lrrrrrrrrrrrr@{}}
\toprule
$1-\alpha$ & \multicolumn{3}{c}{$0.99$} & \multicolumn{3}{c}{$0.95$} & \multicolumn{3}{c}{$0.90$} & \multicolumn{3}{c}{$0.80$} \\
\cmidrule(lr){2-4} \cmidrule(lr){5-7} \cmidrule(lr){8-10} \cmidrule(lr){11-13}
 & Coverage & \multicolumn{2}{c}{Width} & Coverage & \multicolumn{2}{c}{Width} & Coverage & \multicolumn{2}{c}{Width} & Coverage & \multicolumn{2}{c}{Width} \\
Method &  & Mean & Median &  & Mean & Median &  & Mean & Median &  & Mean & Median \\
\midrule
Homo. \ac{nll} & 0.925 & 0.94 & 0.94 & 0.906 & 1.36 & 1.36 & 0.890 & 1.58 & 1.58 & 0.858 & 1.82 & 1.82 \\
Naïve \ac{nll} & 0.980 & 0.96 & 0.75 & 0.958 & 1.38 & 1.09 & 0.937 & 1.58 & 1.26 & 0.893 & 1.83 & 1.45 \\
Nat. \ac{nll} & 0.981 & 0.90 & 0.76 & 0.960 & 1.29 & 1.09 & 0.940 & 1.49 & 1.25 & 0.898 & 1.72 & 1.45 \\
$\beta$-\ac{nll} & 0.980 & 0.78 & 0.71 & 0.957 & 1.12 & 1.03 & 0.933 & 1.29 & 1.19 & 0.885 & 1.49 & 1.37 \\
Faithful & 0.911 & 0.97 & 0.91 & 0.895 & 1.43 & 1.35 & 0.880 & 1.66 & 1.56 & 0.848 & 1.92 & 1.81 \\
Post hoc (\ac{ensemble}) & 0.988 & 1.03 & 0.84 & 0.968 & 1.33 & 1.10 & 0.943 & 1.40 & 1.20 & 0.886 & 1.46 & 1.30 \\
Post hoc (\ac{all}) & 0.969 & 0.62 & 0.53 & 0.932 & 0.87 & 0.73 & 0.894 & 1.00 & 0.84 & 0.817 & 1.16 & 0.97 \\
\ac{scp} & 0.990 & 1.00 & 1.00 & 0.949 & 1.00 & 1.00 & 0.899 & 1.00 & 1.00 & 0.802 & 1.00 & 1.00 \\
\ac{lacp} & 0.989 & 1.09 & 0.83 & 0.947 & 1.11 & 0.85 & 0.897 & 1.11 & 0.86 & 0.795 & 1.12 & 0.86 \\
\midrule
Post hoc (\ac{ensemble})$^*$ & 0.99 & 0.966 & 0.807 & 0.95 & 0.977 & 0.862 & 0.90 & 0.986 & 0.893 & 0.80 & 1.022 & 0.943 \\
Post hoc (\ac{all})$^*$ & 0.99 & 0.858 & 0.723 & 0.95 & 0.964 & 0.808 & 0.90 & 1.021 & 0.854 & 0.80 & 1.100 & 0.920 \\
\bottomrule
\end{tabular}

    }
\end{table}
\begin{table}[h]
    \centering
    \caption{Average rank of methods across all 13 targets (ten \textsc{qm}{\footnotesize 9}, \textsc{utkface}, \textsc{california}, and \textsc{year}) according to their interval score at different nominal levels ($1-\alpha$).}
    \begin{tabular}{lccccc}
\toprule
 & $0.99$ & $0.95$ & $0.90$ & $0.80$ & Mean \\
\midrule
Post hoc (\ac{all}) & 3.0 & 1.6 & 1.7 & 1.8 & 2.02 \\
$\beta$-\ac{nll} & 3.8 & 2.8 & 2.9 & 3.2 & 3.19 \\
Post hoc (\ac{ensemble}) & 2.8 & 3.8 & 4.0 & 3.8 & 3.60 \\
\ac{lacp} & 4.0 & 4.0 & 3.5 & 3.4 & 3.71 \\
\ac{scp} & 5.2 & 5.2 & 5.1 & 4.8 & 5.08 \\
Naïve \ac{nll} & 6.0 & 5.8 & 6.1 & 6.2 & 6.02 \\
Nat. \ac{nll} & 5.8 & 6.2 & 6.2 & 6.1 & 6.04 \\
Homo. \ac{nll} & 6.8 & 7.5 & 7.6 & 7.8 & 7.44 \\
Faithful & 7.5 & 8.1 & 8.0 & 8.0 & 7.90 \\
\bottomrule
\end{tabular}

    \label{tab:intervals}
\end{table}

\FloatBarrier
\section{On Differences to Jimenez \& Katfuzz 2025b}
\FloatBarrier
\label{app:diff_vecchia}

Here we elaborate on the differences to \cite{jimenez2025vecchiagaussianprocessensembles}. In their work, the authors propose using Vecchia Gaussian Process ensembles on latent representations of neural networks trained for mean estimation only in regression tasks, using latent representation collapse as their motivation. Their method is post hoc, and relies on using GPs on all the latent representations throughout a neural network to predict both mean and uncertainty. The authors use training points as conditioning points for all the GPs, and due to the cubic cost of predicting with GPs, use Vecchia conditioning sets to reduce computational complexity. Vecchia conditioning sets are found at inference time by computing $k$ nearest neighbors (where $k$ is a hyperparameter) amongst the training data points for each latent representation $\{z_1,...,z_L\}$. In other words, at inference time their method requires computing $k$ nearest neighbors $L$ times and then conditioning $L$ GPs on these conditioning sets, which is cubic in the number of conditioning points. These $L$ GP predictive posteriors are then combined using a weighted product of predictive densities, where the weighting strategy is an additional hyperparameter choice. This adds up to 3 additional hyperparameter choices that must be made. Finally, their method relies on storing \textit{all} training data such that it can be used at inference time for conditioning, which is not possible for large scale neural networks. The authors mention that using inducing-point GPs or a validation set in place of the training set could remedy this, but never investigate it (and again this would introduce another hyperparameter).

To conclude, we argue that whilst \cite{jimenez2025vecchiagaussianprocessensembles} were the first to use pretrained intermediate representations for UQ in regression, the additional hyperparameter choices makes their method significantly more difficult to use in practice than ours, and that the inference time and storage costs incurred by their method limits the applicability of their method in very large scale neural networks - especially in applications such as molecular dynamics where the model often must be called $500,000$ to $1,000,000$ times for a 1-nanosecond simulation. Moreover, it is not clear how their method would be used on graph neural networks for molecular property prediction, where features are obtained per atom. In contrast, our method simply becomes a sum of the variance contributions computed for each atom as outlined in Eq.~\ref{eq:graph_var}. At a minimum, their method would require significant changes to fit the setting of large scale neural networks and molecular property prediction.

\section{Compute Resources}
Our experiments were completed on various consumer GPUs on an internal cluster, with VRAM in the range of 11-48GB - we did not control which experiments were allocated to which GPU, except for the OMol25 experiments, which we ran on GPUs with a minimum of 24GB VRAM. The storage on these devices were all in the range 256-512GB. Training with each method for a single seed for each of the targets in \cref{tab:qm9_results} took approximately 16 GPU hours, with the exception of our method which can be done on CPUs at very small costs ($\sim$1 CPU hour) if the intermediate representations are cached and saved for later post-hoc fitting. Producing all of the results in \cref{tab:qm9_results} is thus estimated to have taken 4.368 GPU hours. Running the ablation studies can all be done on CPUs as they do not require computing full gradients of large networks, and we estimate these to have taken approximately 1 CPU core hour each.

Running the OMol25 experiments can also be done cheaply by forward propping the hold-out datasets through UMA and AllScAIP once and storing the latent representations. Fitting the post-hoc methods and running tests after this requires negligible CPU hours (can be done in approximately 1 CPU core hour). Forward propping our validation and test sets once however, takes approximately 12 GPU hours for each of the models, leading to a total of approximately 72 GPU hours for the OMol25 experiments.

We thus estimate that our total GPU hour cost was approximately 4500 GPU hours when we include failed runs. All experiments we ran to completion have been included in the final paper.


\newpage
\phantomsection
\section*{NeurIPS Paper Checklist}
\addcontentsline{toc}{section}{NeurIPS Paper Checklist}
\begin{enumerate}

\item {\bf Claims}
    \item[] Question: Do the main claims made in the abstract and introduction accurately reflect the paper's contributions and scope?
    \item[] Answer: \answerYes{} 
    \item[] Justification: The experiments show that our method performs at least on par with existing methods across 13 targets. We provide explicit justification for claims around our method, in our Method \& Experiment Section.

\item {\bf Limitations}
    \item[] Question: Does the paper discuss the limitations of the work performed by the authors?
    \item[] Answer: \answerYes{} 
    \item[] Justification: We provide an explicit limitation section in our Discussion.

\item {\bf Theory assumptions and proofs}
    \item[] Question: For each theoretical result, does the paper provide the full set of assumptions and a complete (and correct) proof?
    \item[] Answer: \answerYes{} 
    \item[] Justification: We do not have theoretical proofs or theoretical results, but we do consider specific model and data conditions that are necessary for the pitfalls we discuss in Section 2 to occur. We explicitly identify the conditions required for variance overfitting, and failure of generalisation of variance predictions. This in particular, is overparametrised neural networks, that are expensive to train.

    \item {\bf Experimental result reproducibility}
    \item[] Question: Does the paper fully disclose all the information needed to reproduce the main experimental results of the paper to the extent that it affects the main claims and/or conclusions of the paper (regardless of whether the code and data are provided or not)?
    \item[] Answer: \answerYes{} 
    \item[] Justification: We provide full experimental setup in the Appendix, and all used data and models are publicly available. On potential acceptance we will release all necessary code.

\item {\bf Open access to data and code}
    \item[] Question: Does the paper provide open access to the data and code, with sufficient instructions to faithfully reproduce the main experimental results, as described in supplemental material?
    \item[] Answer: \answerYes{} 
    \item[] Justification: All used datasets are publicly available, and code will be released upon acceptance, including how to run the experiments and environment setup.

\item {\bf Experimental setting/details}
    \item[] Question: Does the paper specify all the training and test details (e.g., data splits, hyperparameters, how they were chosen, type of optimizer) necessary to understand the results?
    \item[] Answer: \answerYes{} 
    \item[] Justification: All experimental detail and protocols are described in Section 4 and in the Appendix.

\item {\bf Experiment statistical significance}
    \item[] Question: Does the paper report error bars suitably and correctly defined or other appropriate information about the statistical significance of the experiments?
    \item[] Answer: \answerYes{} 
    \item[] Justification: All results in plots and tables are reported with 1 standard error of the mean across random seeds. This is also stated in the table and figure captions.

\item {\bf Experiments compute resources}
    \item[] Question: For each experiment, does the paper provide sufficient information on the computer resources (type of compute workers, memory, time of execution) needed to reproduce the experiments?
    \item[] Answer: \answerYes{} 
    \item[] Justification: We include a description of experiment compute resources in the Appendix.
    
\item {\bf Code of ethics}
    \item[] Question: Does the research conducted in the paper conform, in every respect, with the NeurIPS Code of Ethics \url{https://neurips.cc/public/EthicsGuidelines}?
    \item[] Answer: \answerYes{} 
    \item[] Justification: We do not conflict with anything in the Code of Ethics.

\item {\bf Broader impacts}
    \item[] Question: Does the paper discuss both potential positive societal impacts and negative societal impacts of the work performed?
    \item[] Answer: \answerYes{} 
    \item[] Justification: We include an impact statement.
    
\item {\bf Safeguards}
    \item[] Question: Does the paper describe safeguards that have been put in place for responsible release of data or models that have a high risk for misuse (e.g., pre-trained language models, image generators, or scraped datasets)?
    \item[] Answer: \answerNA{} 
    \item[] Justification: There are no high risk misuses directly connected to our work.

\item {\bf Licenses for existing assets}
    \item[] Question: Are the creators or original owners of assets (e.g., code, data, models), used in the paper, properly credited and are the license and terms of use explicitly mentioned and properly respected?
    \item[] Answer: \answerYes{} 
    \item[] Justification: We cite all used datasets, methods and pretrained models in our work. However, we do not explicitly list the license and terms of use for each dataset.

\item {\bf New assets}
    \item[] Question: Are new assets introduced in the paper well documented and is the documentation provided alongside the assets?
    \item[] Answer: \answerNA{} 
    \item[] Justification: We do not release new assets as part of this work.

\item {\bf Crowdsourcing and research with human subjects}
    \item[] Question: For crowdsourcing experiments and research with human subjects, does the paper include the full text of instructions given to participants and screenshots, if applicable, as well as details about compensation (if any)? 
    \item[] Answer: \answerNA{} 
    \item[] Justification: Our paper does not involve crowdsourcing nor research with human subjects.

\item {\bf Institutional review board (IRB) approvals or equivalent for research with human subjects}
    \item[] Question: Does the paper describe potential risks incurred by study participants, whether such risks were disclosed to the subjects, and whether Institutional Review Board (IRB) approvals (or an equivalent approval/review based on the requirements of your country or institution) were obtained?
    \item[] Answer: \answerNA{} 
    \item[] Justification: Our paper does not involve crowdsourcing nor research with human subjects.

\item {\bf Declaration of LLM usage}
    \item[] Question: Does the paper describe the usage of LLMs if it is an important, original, or non-standard component of the core methods in this research? Note that if the LLM is used only for writing, editing, or formatting purposes and does \emph{not} impact the core methodology, scientific rigor, or originality of the research, declaration is not required.
    \item[] Answer: \answerNA{} 
    \item[] Justification: The core method development in this research does not involve LLMs as any important, original, or non-standard components.

\end{enumerate}

\end{document}